\documentclass[10pt,twocolumn,letterpaper]{article}

\usepackage{cvpr}              

\usepackage[dvipsnames]{xcolor}

\usepackage{multirow}
\usepackage{amsmath}
\usepackage{stfloats} 
\usepackage{tikz}
\usetikzlibrary{shapes.geometric, arrows}
\usetikzlibrary{positioning}
\usepackage{float}
\usepackage{tabularx}
\usepackage[accsupp]{axessibility}

\definecolor{cvprblue}{rgb}{0.21,0.49,0.74}
\definecolor{midday_blue}{RGB}{0,169,206}
\usepackage[pagebackref,breaklinks,colorlinks,citecolor=cvprblue]{hyperref}

\usepackage{multirow}
\usepackage{color, colortbl}
\usepackage{xcolor}
\usepackage[dvipsnames]{xcolor}
\definecolor{gray}{gray}{0.85}

\def\paperID{7} 
\def\confName{CVPR}
\def\confYear{2024}

\title{Orientation-conditioned Facial Texture Mapping for Video-based Facial Remote Photoplethysmography Estimation}

\author{
    Sam Cantrill$^{1,2}$\thanks{Denotes corresponding author.}\quad 
    David Ahmedt-Aristizabal$^{2}$ \quad \\
    Lars Petersson$^{2}$ \quad
    Hanna Suominen$^{1,3}$ \quad 
    Mohammad Ali Armin$^{2}$ \\
    {$^{1}$Australian National University, Canberra, Australia} \\
    {$^{2}$Data61, Commonwealth and Scientific Industrial Research Organization, Canberra, Australia} \\
    {$^{3}$University of Turku, Turku, Finland} \\
    {\tt\small\{sam.cantrill, hanna.suominen\}@anu.edu.au} \\
    {\tt\small\{ali.armin, david.ahmedtaristizabal, lars.petersson\}@data61.csiro.au}
}

\begin{document}
\twocolumn[{%
\renewcommand\twocolumn[1][]{#1}%
\maketitle
\vspace{-24pt}
\begin{center}
    \centering
    \captionsetup{type=figure}

    \begin{tikzpicture}[node distance=0.20cm and 0.50cm]
        \fill [fill=white, opacity=0] (0,0) rectangle (0.95\linewidth,-4.50);

        \fill [fill=gray, opacity=0.05, rounded corners=5] (-1.55,1.25) rectangle (1.55,-3.70);
        \draw [gray!50!black, thick, opacity=0.25, rounded corners=5] (-1.55,1.25) rectangle 
        (1.55,-3.70);
        \node [below, align=center] at (0, -3.70) {Video};

        \fill [fill=midday_blue, opacity=0.075, rounded corners=5] (1.75,1.25) rectangle (10.40,-3.70);
        \draw [midday_blue!50!black, thick, opacity=0.45, rounded corners=5] (1.75,1.25) rectangle (10.40,-3.70);
        \node [below, align=center] at (6.075, -3.70) {Orientation-conditioned Facial Texture \\Video Pipeline};

        \fill [fill=gray, opacity=0.05, rounded corners=5] (10.60,1.25) rectangle (13.15,-3.70);
        \draw [gray!50!black, thick, opacity=0.25, rounded corners=5] (10.60,1.25) rectangle (13.15,-3.70);
        \node [below, align=center] at (11.875, -3.70) {Video-based \\Model};

        \fill [fill=gray, opacity=0.05, rounded corners=5] (13.35,1.25) rectangle (15.90,-3.70);
        \draw [gray!50!black, thick, opacity=0.25, rounded corners=5] (15.90,1.25) rectangle (13.35,-3.70);
        \node [below, align=center] at (14.625, -3.70) {rPPG + Pulse Rate \\Estimation};

        \node (xy_frame) at (0,0) {\includegraphics[height=2.0cm]{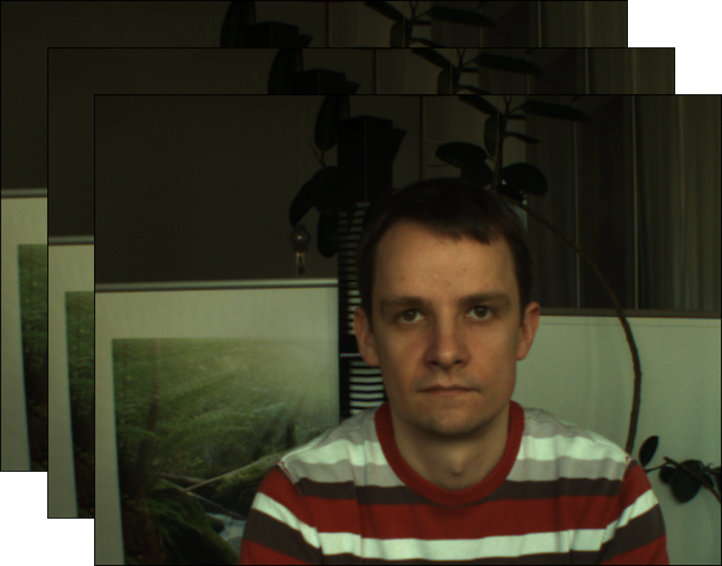}};

        \node (xy_frame_overlay) [right=of xy_frame] {\includegraphics[height=2.0cm]{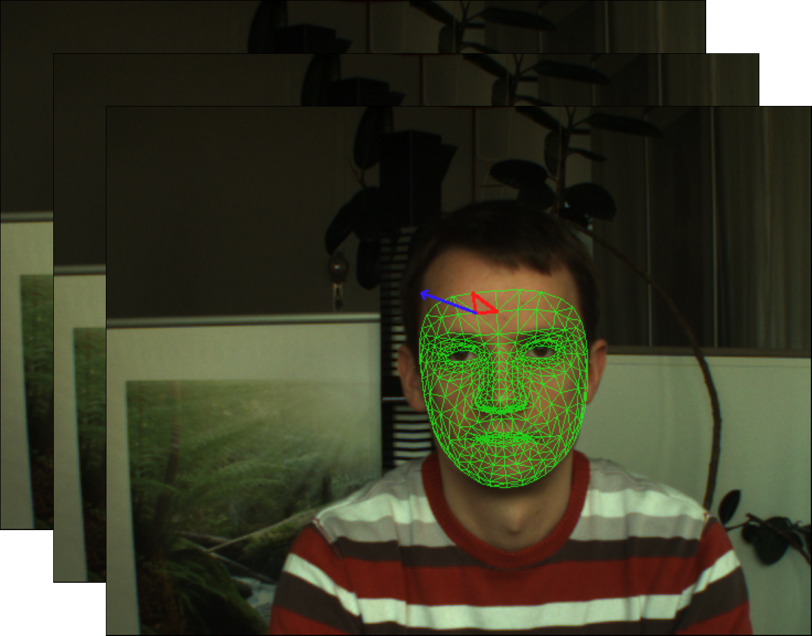}};
        \node (uv_frame_overlay) [right=of xy_frame_overlay] {\includegraphics[height=2.0cm]{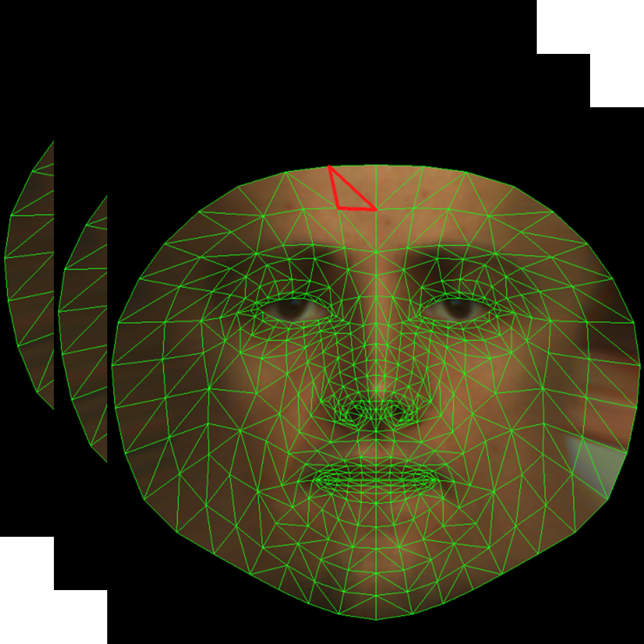}};
        \node (uv_masked_frame) [right=of uv_frame_overlay] {\includegraphics[height=2.0cm]{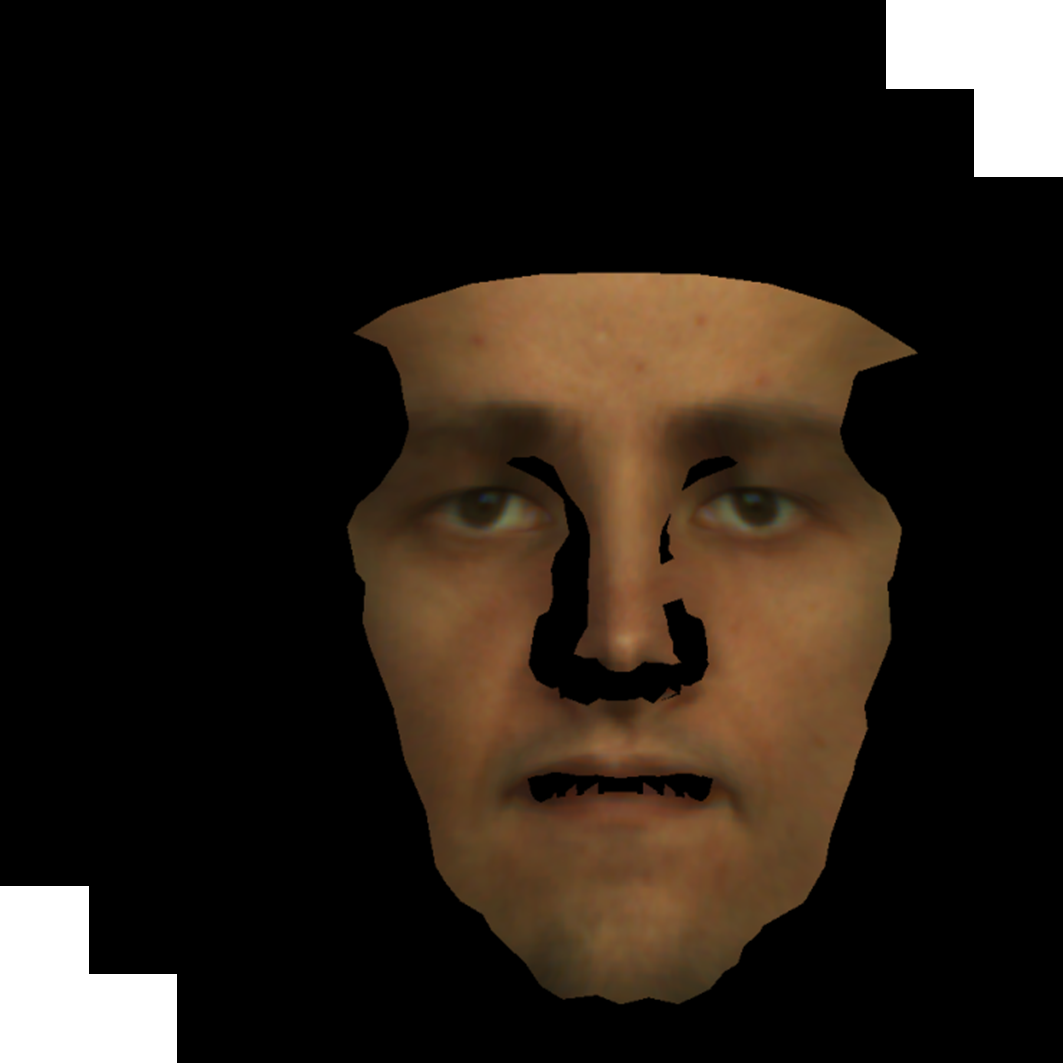}};

        \node (uv_angle) [below=of uv_frame_overlay, xshift=-1.50cm] {\includegraphics[height=2.0cm]{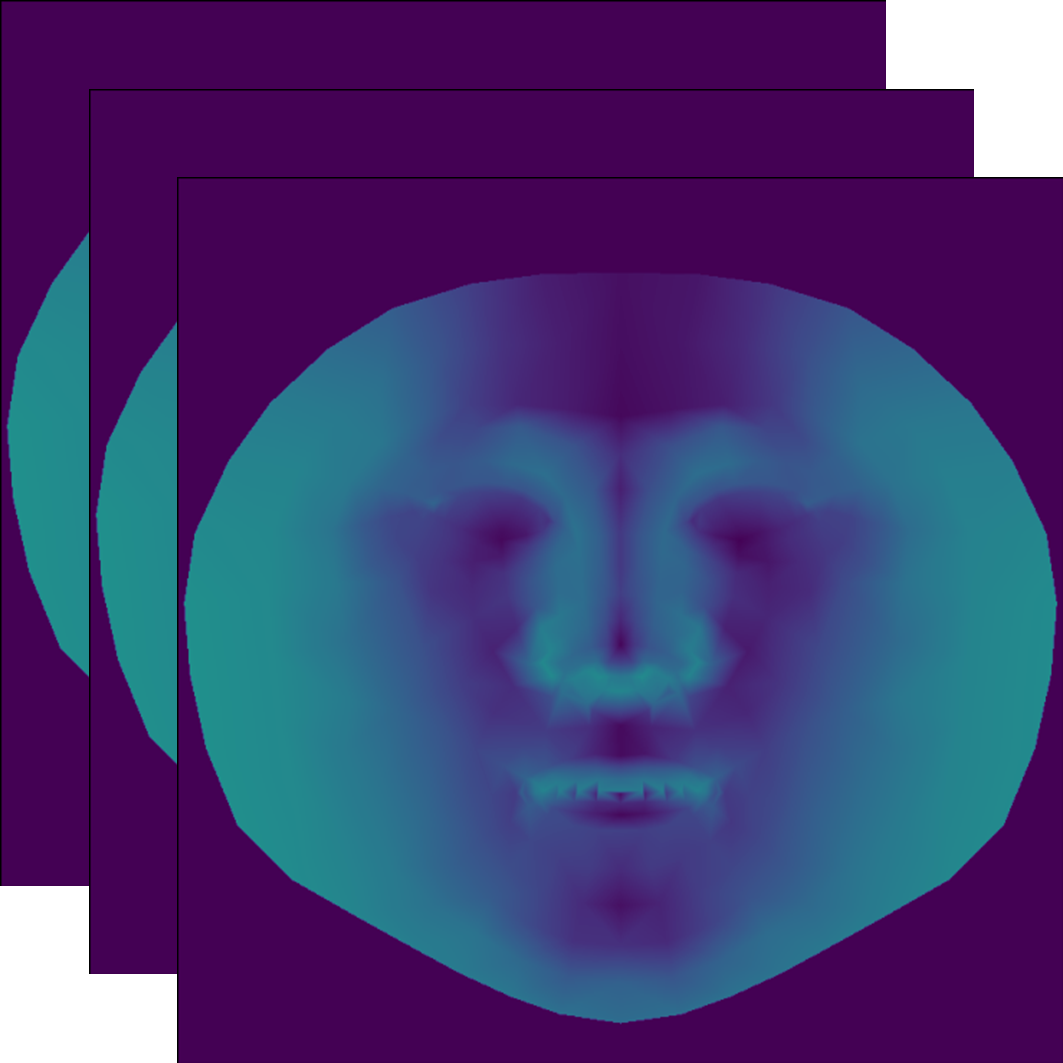}};
        \node (uv_mask) [right=of uv_angle, xshift=0.13cm] {\includegraphics[height=2.0cm]{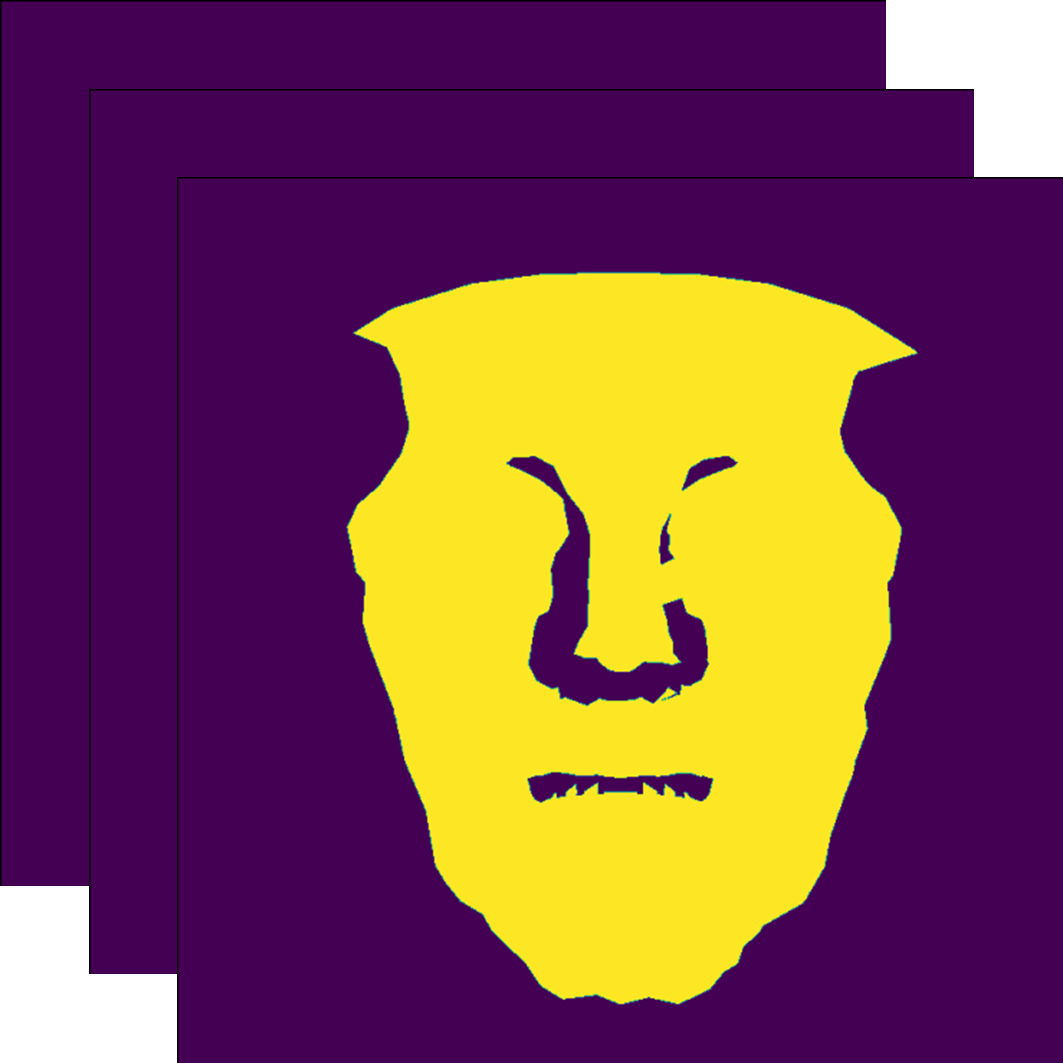}};

        \node (model_node) [fill opacity=0.00, minimum width=2.26cm, minimum height=2.26cm, right=of uv_masked_frame] {};
        \node [draw, fill=white, fill opacity=1.00, rounded corners=5, minimum width=2.0cm, minimum height=2.0cm, inner sep=-10.0cm] (model) at (model_node) {};
        \node [xshift=-0.25cm] (model) at (model_node) {\includegraphics[height=1.75cm]{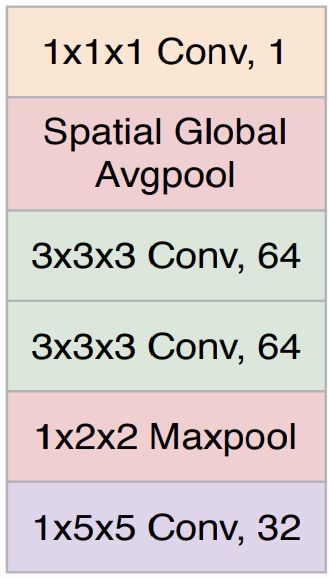}};
        \node [rotate=90, align=center, xshift=0cm, yshift=-0.50cm, font=\fontsize{7}{10}\selectfont] at (model_node) {PhysNet-3DCNN};

        \node (rppg_signal_node) [fill opacity=0.00, minimum width=2.26cm, minimum height=2.26cm, right=of model_node] {};
        \node [draw, fill=white, fill opacity=1.00, rounded corners=5, minimum width=2cm, minimum height=2cm] (rppg_signal_box) at (rppg_signal_node) {};
        \node (rppg_signal) at (rppg_signal_node) {\includegraphics[height=1.75cm]{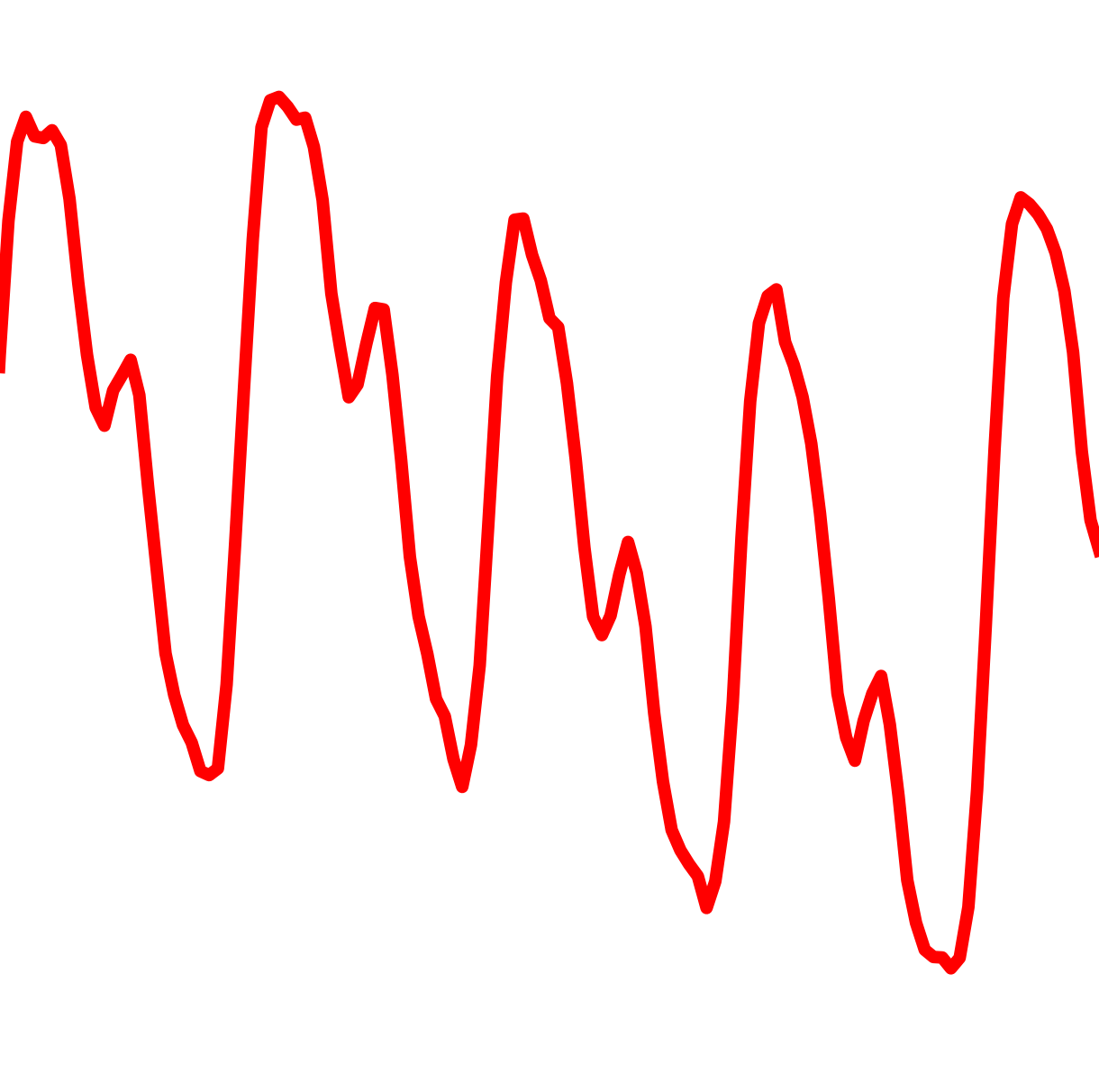}};

        \node (evaluation_node) [fill opacity=0.00, minimum width=2.26cm, minimum height=2.26cm, below=of rppg_signal_node] {};
        \node [draw, fill=white, fill opacity=1.00, rounded corners=5, minimum width=2cm, minimum height=2cm] (evaluation) at (evaluation_node) {};
        \node [inner sep=0pt] (freq) at (evaluation_node) {\includegraphics[height=1.75cm]{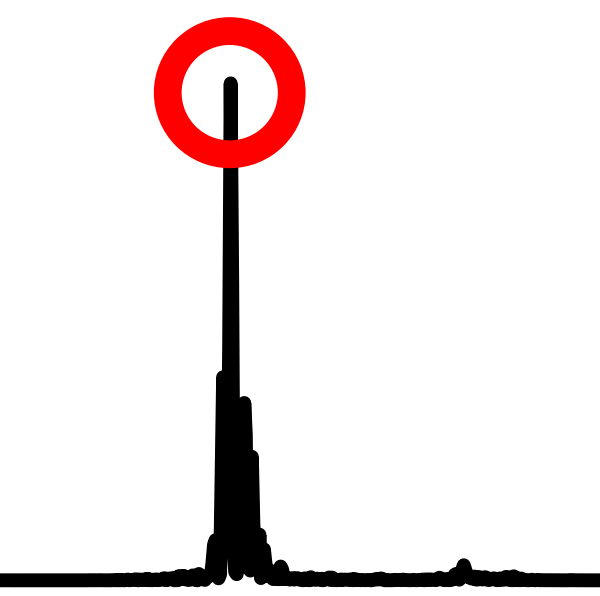}};
        \node [inner sep=0pt, xshift=0.50cm, yshift=0.50cm] (heart) at (evaluation_node) {\includegraphics[height=0.75cm]{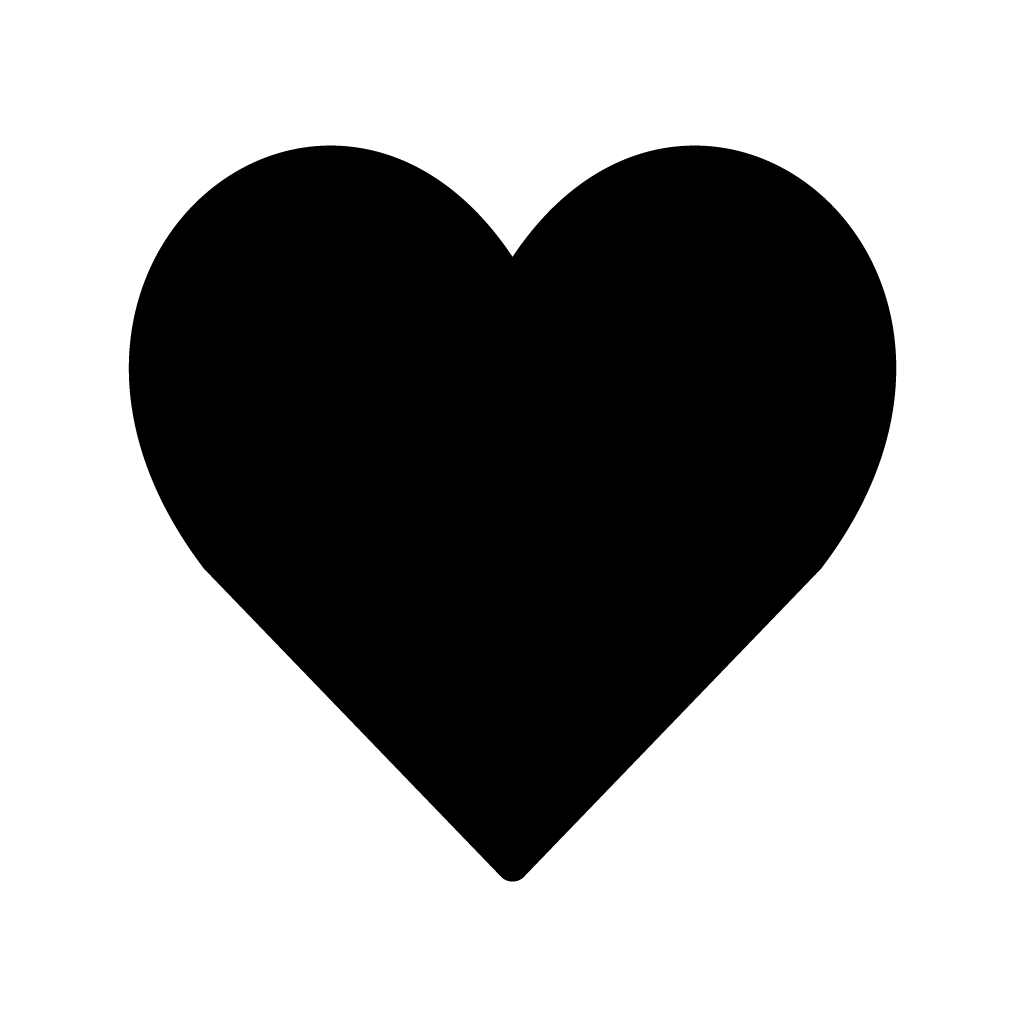}};

        \draw[->, line width=1.0pt] (xy_frame) -- (xy_frame_overlay);
        \draw[->, line width=1.0pt] (xy_frame_overlay) -- (uv_frame_overlay);
        \draw[->, line width=1.0pt] (uv_frame_overlay) -- (uv_masked_frame);
        \draw[->, line width=1.0pt] (uv_masked_frame) -- (model_node);
        \draw[->, line width=1.0pt] (model_node) -- (rppg_signal_node);
        \draw[->, line width=1.0pt] (rppg_signal_node) -- (evaluation_node);
        \draw[->, line width=1.0pt] (xy_frame_overlay) |- (uv_angle);
        \draw[->, line width=1.0pt] (uv_angle) -- (uv_mask);
        \draw[->, line width=1.0pt] (uv_mask.north) -- (7.72,-0.05);

    \end{tikzpicture}
    
    \vspace{-9pt}
    \captionof{figure}{
    Proposed methodology for constructing the orientation-conditioned facial texture video using UV-coordinate texture mapping to enhance the motion robustness of camera-based remote photoplethysmography (rPPG) and downstream pulse rate (PR) estimation.
    } 
    \label{fig:pipeline_summary}
\end{center}
}]
\vspace{+8pt}

\begin{abstract}
\vspace{-8pt} 
Camera-based remote photoplethysmography (rPPG) enables contactless measurement of important physiological signals such as pulse rate (PR).
However, dynamic and unconstrained subject motion introduces significant variability into the facial appearance in video, confounding the ability of video-based methods to accurately extract the rPPG signal.
%
%
In this study, we leverage the 3D facial surface to construct a novel orientation-conditioned facial texture video representation which improves the motion robustness of existing video-based facial rPPG estimation methods.
Our proposed method achieves a significant 18.2\% performance improvement in cross-dataset testing on MMPD over our baseline using the PhysNet model trained on PURE, highlighting the efficacy and generalization benefits of our designed video representation.
%
%
We demonstrate significant performance improvements of up to 29.6\% in all tested motion scenarios in cross-dataset testing on MMPD, even in the presence of dynamic and unconstrained subject motion, emphasizing the benefits of disentangling motion through modeling the 3D facial surface for motion robust facial rPPG estimation.
We validate the efficacy of our design decisions and the impact of different video processing steps through an ablation study.
Our findings illustrate the potential strengths of exploiting the 3D facial surface as a general strategy for addressing dynamic and unconstrained subject motion in videos.
The code is available at \href{https://samcantrill.github.io/orientation-uv-rppg/}{https://samcantrill.github.io/orientation-uv-rppg/}.

\end{abstract}
\vspace{-8pt}    
\section{Introduction}
\label{sec:intro}

In camera-based remote photoplethysmography (rPPG) we estimate the rPPG signal using video obtained from consumer-grade cameras. The rPPG signal contains valuable and meaningful physiological information including pulse rate (PR), respiration rate (RR), and pulse rate variability (PRV)~\cite{camerameasurementphysiological-mcduff-2023}. 
Contactless approaches offer distinct advantages over traditional contact-based methods, consequently finding applications in telehealth~\cite{multitasktemporalshift-liu-2021}, in-patient monitoring~\cite{noncontactheartrate-aarts-2013a}, and various affective computing tasks~\cite{multimodalspontaneousemotion-zhang-2016, facelivenessdetection-lin-2019, deepfakesonphysdeepfakesdetection-hernandez-ortega-2020, transrppgremotephotoplethysmography-yu-2021}.
Despite significant advancement in facial rPPG estimation, existing methods often display performance degradation in challenging real-world scenarios~\cite{realtimewebcamheartrate-gudi-2020, multitasktemporalshift-liu-2021, mmpdmultidomainmobile-tang-2023}, particularly in handling unconstrained and dynamic subject motion, thus limiting the potential of this technology.



Facial rPPG estimation methods primarily rely on detecting subtle color changes on the face surface caused by the sub-surface blood volume pulse (BVP) to estimate downstream physiological signals such as pulse rate (PR)~\cite{noncontactautomatedcardiac-poh-2010, robustpulserate-dehaan-2013, algorithmicprinciplesremote-wang-2017}. 
However, unconstrained and dynamic subject motion introduces significant variability into the observed appearance of the face in video, with large pixel-level variations potentially overshadowing the subtle changes associated with the rPPG signal.



Continued advancement in facial rPPG estimation has solidified the strengths of deep-learning based methods~\cite{visualheartrate-spetlik-2018, deepphysvideobasedphysiological-chen-2018, synrhythmlearningdeep-niu-2018, multitasktemporalshift-liu-2021} in capturing the intricate and non-linear relationships between spatio-temporal input features in video and the target rPPG signal. 
Video-based approaches~\cite{remotephotoplethysmographsignal-yu-2019, physformerfacialvideobased-yu-2022, efficientphysenablingsimple-liu-2023} operate directly on video-formatted data, demonstrating robust spatio-temporal modeling capabilities. However, they are easily influenced by real-world conditions~\cite{autohrstrongendtoend-yu-2020} such as subject motion.
Spatio-temporal map (STMap)-based methods~\cite{synrhythmlearningdeep-niu-2018, rhythmnetendtoendheart-niu-2020, dualganjointbvp-lu-2021, rppgmaeselfsupervisedpretraining-liu-2023} leverage prior knowledge about the characteristics of the subtle rPPG signal to design hand-crafted 2D spatio-temporal input representations.
Consequently, STMap-based methods often exhibit superior performance and robustness over video-based approaches. However, their coarse spatial representation limits the extraction of rich and contextual spatio-temporal features.





Modeling and localizing the spatial region of the face in a video through facial detection represents a strong inductive bias for facial rPPG estimation. It is frequently used to enhance performance, rendering it a standard component of the input processing pipeline~\cite{rppgtoolboxdeepremote-liu-2023} across state-of-the-art video-based~\cite{physformerfacialvideobased-yu-2022} and STMap-based~\cite{rppgmaeselfsupervisedpretraining-liu-2023} approaches. 
A limited number of methods~\cite{face2ppgunsupervisedpipeline-casado-2023, dualpathtokenlearnerremote-qian-2023} have leveraged facial structure modeling to extract dynamic surface regions instead of fixed spatial regions from video, aiming to improve robustness against subject translation, rotation, and facial expressions. 
Several studies have noted the importance of considering the 3D facial structure~\cite{optimisingrppgsignal-wong-2022, remoteheartrate-maki-2020} for improving both performance and motion robustness.
However, within video-based methods, there is a notable absence of works that exploit facial structure for enhancing the motion robustness of rPPG estimation.




In this work, we investigate how modeling the 3D facial structure can be used as a fundamental strategy for disentangling rigid and non-rigid subject motion from video. 
%
%
UV coordinate texture maps are commonly used to represent the texture of a 3D surface using an image, providing a way to represent the 3D surface of the face in a form compatible with existing video-based facial rPPG estimation methods, as illustrated in \Cref{fig:pipeline_summary}.
Our methodology uses existing 3D landmark detection techniques to model the 3D surface of the face as a mesh. Subsequently, we apply UV coordinate texture mapping to warp the observed facial surface within video frames into a facial texture representation. 
%
%
Given the introduction of geometric distortion during the transformation, we mask the UV coordinate facial texture using the facial surface orientation to remove regions with re-projected and distorted appearance. 
We demonstrate a significant improvement in generalization performance across all motion scenarios through cross-dataset testing, employing a well-understood baseline method.

Our contributions are summarized as~follows:
\begin{itemize}
    \item We propose a novel UV-coordinate facial texture video representation conditioned on facial surface orientation designed to be compatible with video-based methods which significantly enhances the generalization performance and motion-robustness of facial rPPG estimation.

\end{itemize}
 
\section{Related Work}
\label{sec:related}

\noindent\textbf{Signal Processing Methods:}
Early efforts in camera-based facial rPPG estimation primarily focused on extracting temporal signals with domain-specific knowledge such as blind source separation (BSS) methods and maximum periodicity criteria~\cite{noncontactautomatedcardiac-poh-2010, advancementsnoncontactmultiparameter-poh-2011}. These approaches are restricted to limited motion conditions, under which these assumptions remain valid. Subsequent works sought to improve the robustness of the rPPG signal extraction against sources of noise such as subject motion and illumination by exploiting prior domain knowledge. For instance, leveraging knowledge about skin optical properties~\cite{robustpulserate-dehaan-2013} or the blood volume pulse dynamics~\cite{improvedmotionrobustness-dehaan-2014} helped isolate a more robust color vector sub-space with a higher signal-to-noise ratio. Further works like S2R~\cite{novelalgorithmremote-wang-2016} and POS~\cite{algorithmicprinciplesremote-wang-2017} adopted data-driven approaches to dynamically extract such a sub-space. 

Despite the strong priors these methods employ, they suffer from significant performance degradation in the presence of unconstrained and dynamic subject motion, as the underlying assumptions may not hold. As a result, deep-learning methods gained prominence due to their ability to learn the complex and non-linear relationship between the spatio-temporal video features and the rPPG signal.





\noindent\textbf{STMap-based Deep Learning Methods:}
Spatial-temporal map (STMap)-based methods in facial rPPG estimation rely on hand-crafted spatial-temporal representations of video to exploit prior knowledge about the rPPG signal, aiming to improve the signal-to-noise ratio. Similarly to signal processing methods, they leverage information about the facial position and surface within video alongside prior domain knowledge as strong inductive~biases.

Construction of the STMap or multi-scale STMap (MSTMap) commonly involves extracting a static spatial region of interest from video containing the face~\cite{synrhythmlearningdeep-niu-2018, videobasedremotephysiological-niu-2020, dualganjointbvp-lu-2021, rppgmaeselfsupervisedpretraining-liu-2023}, dividing it into grid-cells, and processing them into STMap-pixels for each time-step. While this method aids the performance by increasing the signal-to-noise ratio of the rPPG signal in video, it remains susceptible to both subject translation, head rotation and facial expressions. Other approaches extract dynamic 2D facial surface regions of interest~\cite{dualpathtokenlearnerremote-qian-2023}, providing improved robustness to subject translation and subject rotation, as each STMap pixel represents a temporally consistent region on the face surface. 

Despite the advantages of disentangling subject motion through the input representation, it has not been explored whether using the dynamic 3D surface of the face can be used to further improve the robustness to motion. Furthermore, the coarse nature of STMap pixel-regions limits the ability to extract the rich spatio-temporal information present in video~\cite{benefitdistractiondenoising-nowara-2021,transrppgremotephotoplethysmography-yu-2021}.

\noindent\textbf{Video-based Deep Learning Methods:}
Video-based methods differ from STMap-based methods by directly operating on video-formatted data, sequences of spatial frames, employing models with strong spatio-temporal modeling capabilities to learn to extract subtle changes in skin color due to the rPPG signal. The performance and robustness of video-based methods implicitly rely on regularization in the learning process. Despite the strengths of video-based methods they are often outperformed by STMap-based methods which employ more explicit inductive biases for facial rPPG estimation~\cite{autohrstrongendtoend-yu-2020}.

Given the relevance of the facial surface as the object of analysis for facial rPPG estimation, numerous video-based methods explore different network architectures and functional forms to refine spatial features and improve motion robustness. Some approaches incorporate explicit spatial masking in feature space \cite{deepphysvideobasedphysiological-chen-2018, hearttrackconvolutionalneural-perepelkina-2020, benefitdistractiondenoising-nowara-2021, multitasktemporalshift-liu-2021} to guide the model's attention to salient spatial regions. Other works utilize spatial feature refinement modules \cite{robustremoteheart-niu-2019, robustheartrate-hu-2022, noncontactppgsignal-li-2023} to allow the model to implicitly learn more informative spatial features. Despite performance improvements, such approaches are still sensitive to various motion scenarios due to a lack of an explicit inductive bias for motion robustness.

Video input representation can also be used to exploit facial and motion features, enhancing motion robustness. DeepPhys~\cite{deepphysvideobasedphysiological-chen-2018} proposed the normalized frame-difference as a motion representation to better guide motion estimation, a technique employed in further works~\cite{multitasktemporalshift-liu-2021, rppgtoolboxdeepremote-liu-2023}. 
Optical flow representations have also been used to provide motion context to the model, guiding the spatial feature alignment over time to improve the robustness to motion~\cite{learningmotionrobustremote-li-2022}. 
Furthermore, leveraging facial position through static facial detection is commonly employed in state-of-the-art approaches~\cite{physformerfacialvideobased-yu-2022} to increase the signal-to-noise ratio of the rPPG signal in the video.

Using different motion-based and frame-based video representations can provide a strong and explicit inductive bias for motion robustness. However, leveraging the 3D surface of the face to disentangle rigid and non-rigid motion has not been explored as a general mechanism to enhance the robustness of facial rPPG estimation.




\section{Method}
\label{sec:method}

\begin{figure}[H]

  \centering
  
  \begin{tikzpicture}[node distance=1.0cm and 1.0cm]

    \node (xy_frame) at (0,0) {\includegraphics[height=2.5cm]{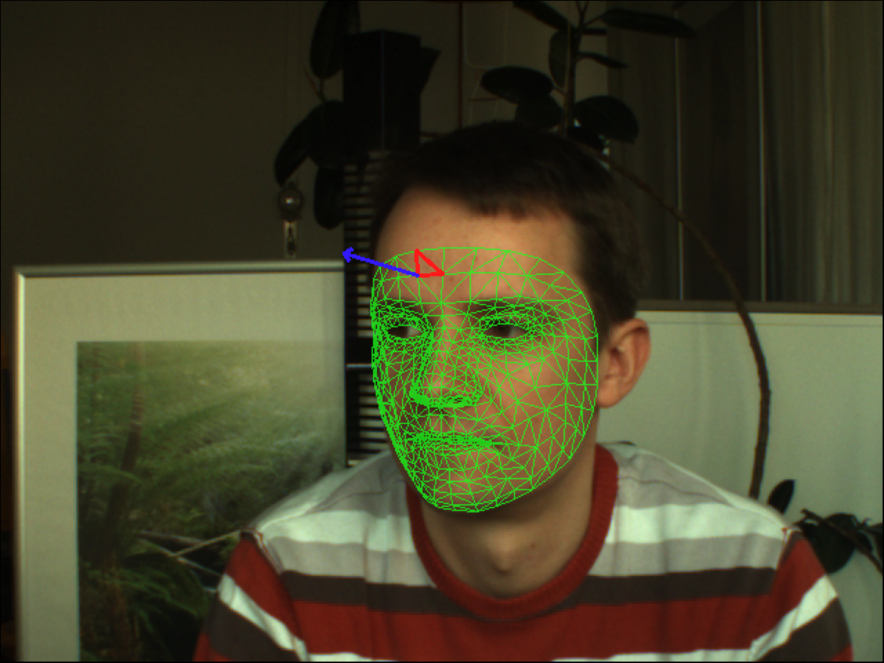}};
    
    \node (uv_frame) [right=of xy_frame, xshift=-1.0cm] {\includegraphics[height=2.5cm]{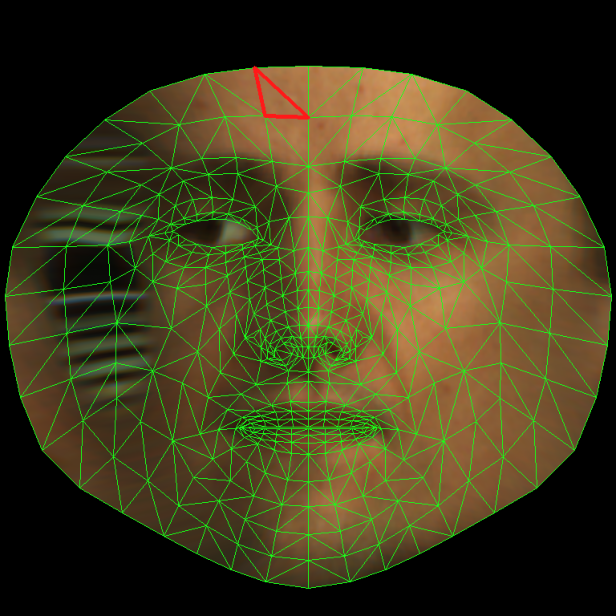}};
    
  \end{tikzpicture}
  
  \caption{Example from PURE~\cite{noncontactvideobasedpulse-stricker-2014} of a XY coordinate image-space frame and the computed UV coordinate texture-space frame with overlaid 3D facial meshes.}%
  \label{fig:xy_and_uv_frame}
\vspace{-8pt}
\end{figure}

\begin{figure*}[t]
 
    \centering

    \begin{tikzpicture}[node distance=1cm and 2cm]

        \fill [fill=midday_blue, opacity=0.075, rounded corners=5] (-1.55,1.40) rectangle (14.90,-4.80);
        \draw [midday_blue!50!black, thick, opacity=0.25, rounded corners=5] (-1.55,1.40) rectangle (14.90,-4.80);
    
        \node (xy_frame) at (0,0) {\includegraphics[height=2.0cm]{media/VIDEO_PURE_01-01_1392643995576148992_XY_frame.png}};
        \node (xy_frame_overlay) [right=of xy_frame] {\includegraphics[height=2.0cm]{media/VIDEO_PURE_01-01_1392643995576148992_XY_frame_overlay.png}};
        \node (uv_frame_overlay) [right=of xy_frame_overlay] {\includegraphics[height=2.0cm]{media/VIDEO_PURE_01-01_1392643995576148992_UV_frame.png}};
        \node (uv_angle) [below=of xy_frame_overlay, xshift=1.50cm] {\includegraphics[height=1.50cm]{media/VIDEO_PURE_01-01_1392643995576148992_UV_angle.png}};
        \node (uv_mask) [right=of uv_angle] {\includegraphics[height=1.50cm]{media/VIDEO_PURE_01-01_1392643995576148992_UV_mask.png}};
        \node (uv_frame_masked) [right=of uv_frame_overlay] {\includegraphics[height=2.0cm]{media/VIDEO_PURE_01-01_1392643995576148992_UV_frame_mask.png}};

        \node (camera) at (2, -2) {\includegraphics[height=0.50cm, angle=-45, scale=-1]{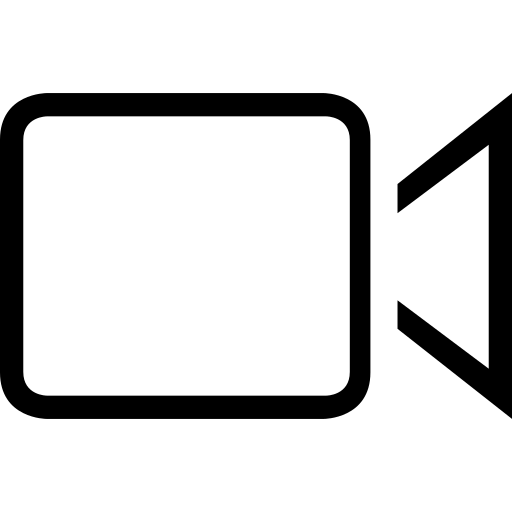}};
        \draw[->, line width=1.0pt] (1.5,-1.5) -- (1.75,-1.75);
        \node [below, xshift=0.50cm, yshift=-0.25cm] at (camera.west) {$\vec{n}_{cam}$};

        \node [below, align=center] at (xy_frame.south) {XY Appearance \\Frames};
        \node [below, align=center] at (xy_frame_overlay.south) {XY Appearance Frames \\\& 3D Meshes};
        \node [below, align=center] at (uv_frame_overlay.south) {UV Appearance \\Frames};
        \node [below, align=center] at (uv_angle.south) {UV Angle \\Frames, $\Theta_{UV}$};
        \node [below, align=center] at (uv_mask.south) {UV Appearance \\Masks};
        \node [below, align=center] at (uv_frame_masked.south) {Orientation-\\conditioned \\Facial Texture \\Frames};
        
        \draw[->, line width=1.0pt] (xy_frame) -- (xy_frame_overlay) node[midway, above, align=center] {3D Landmark \\Detection};
        \draw[->, line width=1.0pt] (xy_frame_overlay) -- (uv_frame_overlay) node[midway, above, align=center] {Frame \\Transform};
        \draw[->, line width=1.0pt] (xy_frame_overlay.south) ++(0,-1.0) |- (uv_angle.west) node[midway, below, align=center, xshift=-1.00cm, yshift=0.00cm] {Compute $\theta_{v}$ \& \\Transform};
        \draw[->, line width=1.0pt] (uv_frame_overlay) -- (uv_frame_masked) node[midway, above, align=center] {Masking};
        \draw[->, line width=1.0pt] (uv_angle) -- (uv_mask) node[midway, above, align=center] {Threshold \\$\Theta_{UV}$};
        \draw[->, line width=1.0pt] (uv_mask) -| (11.50,-0.05);
        
    \end{tikzpicture}

    \caption{Pipeline for constructing orientation-conditioned facial texture video from input video frames. It leverages a temporally coherent 3D facial mesh~\cite{mediapipeframeworkbuilding-lugaresi-2019} to warp the observed XY coordinate facial surface into a pre-defined UV coordinate texture-space~\cite{mediapipeframeworkbuilding-lugaresi-2019}, followed by masking based on orientation, $\Theta_{UV}$, between the camera and the facial surface to reduce appearance distortion.}
    \label{fig:method/pipeline}
\vspace{-10pt}    
\end{figure*}

UV coordinate texture maps are used to represent the texture of a 3D surface using an image as shown in \Cref{fig:xy_and_uv_frame}. UV coordinate texture mapping can therefore be used to disentangle both rigid and non-rigid facial motion from the observed regions of the face within a video frame.
This UV frame representation can be leveraged to reduce motion-related feature variability within a video, thus enhancing the motion robustness of video-based facial rPPG estimation method.
\Cref{fig:method/pipeline} outlines the pipeline for constructing the orientation-conditioned facial texture video representation.

\subsection{UV Facial Texture Representation}
\label{sec:method/input}
We begin by modeling the 3D geometry of the face within the image-plane using the MediaPipe FaceMesh~\cite{mediapipeframeworkbuilding-lugaresi-2019} which detects 468 3D facial landmarks in a non-metrical geometry-space per frame. In this geometry-space, the camera normal is aligned with the negative z-axis, $\vec{n}_{cam} = [0, 0, -1]^{T}$. These landmarks are represented as a mesh using a pre-defined triangular tessellation scheme~\cite{mediapipeframeworkbuilding-lugaresi-2019}.

\noindent\textbf{Computing the UV Frame:}
To compute the facial texture representation, we start by projecting the XYZ coordinate geometry-space facial landmarks onto the XY-plane, yielding the XY coordinate image-space facial landmarks. 
To perform UV coordinate texture mapping, we compute a series of affine transformations based on the mesh triangles in the XY coordinate image-space, aligning them with the pre-defined and corresponding triangles in UV coordinate texture-space~\cite{mediapipeframeworkbuilding-lugaresi-2019}. These affine transformations are then applied to the video frame pixels, converting them from XY coordinate image-space to UV coordinate texture-space. 
We utilize bi-linear interpolation to compute missing values between transformed points, ensuring smooth transitions. We fill undefined areas with 0, as no points exist outside of the facial convex hull in UV coordinate texture-space this naturally provides facial segmentation. Thus we obtain the facial texture representation, an example of this is shown in \Cref{fig:xy_and_uv_frame}.
This method of computing the facial texture representation through inherently provides both dynamic localization and extraction of the facial surface.

\begin{figure}
  \centering
  \begin{tikzpicture}[node distance=1.0cm and 1.0cm]

    \node (uv_frame_90) at (0,0) {\includegraphics[height=1.75cm]{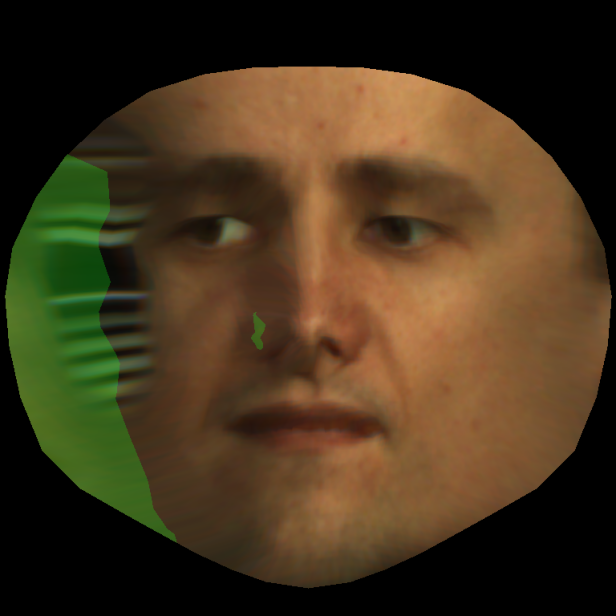}};
    
    \node (uv_frame_60) [right=of uv_frame_90, xshift=-1.0cm] {\includegraphics[height=1.75cm]{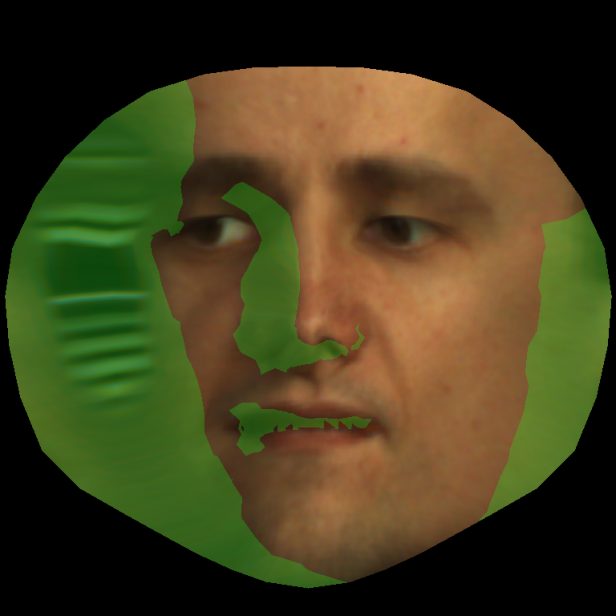}};

    \node (uv_frame_45) [right=of uv_frame_60, xshift=-1.0cm] {\includegraphics[height=1.75cm]{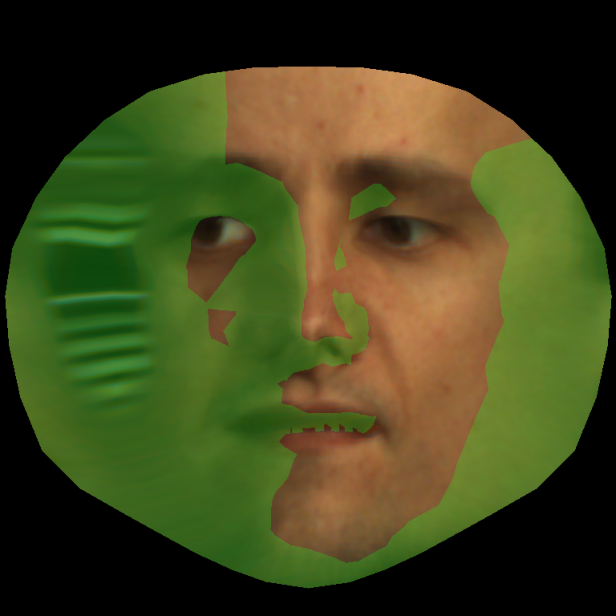}};

    \node (uv_frame_30) [right=of uv_frame_45, xshift=-1.0cm] {\includegraphics[height=1.75cm]{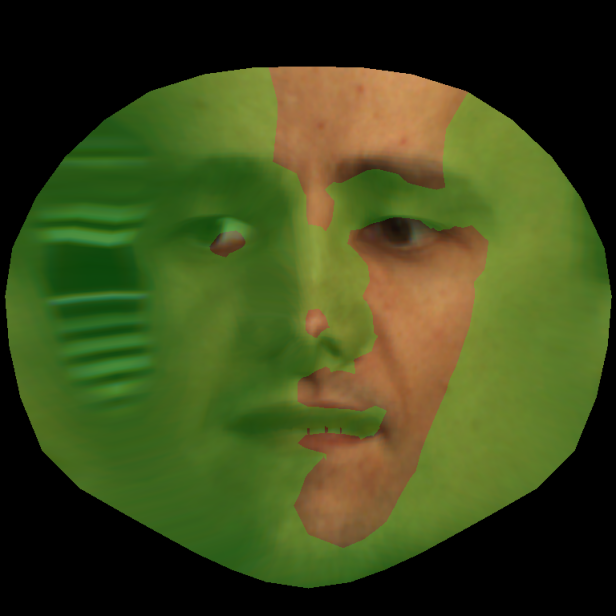}};
    
  \end{tikzpicture}
  
  \caption{Example of a UV texture-space frame from PURE~\cite{noncontactvideobasedpulse-stricker-2014} with facial surface highlighted green based on the relative angle between the surface and the camera of $\Theta_{UV} \geq 90^{\circ}$, $60^{\circ}$, $45^{\circ}$, and $30^{\circ}$ respectively, to highlight regions with re-projected and/or distorted appearance.}
  \label{fig:uv_frame_distortion}
\vspace{-16pt}
\end{figure}

However, we note two issues that arise due to the facial texture mapping transformation. Firstly, facial surface regions with a normal vector obtuse to $\vec{n}_{cam}$ will be re-projected onto the XY plane and contain duplicate appearance information. Secondly, facial surface regions with a normal vector close to perpendicular to $\vec{n}_{cam}$ will experience significant distortion of the appearance when mapping to UV-space. These issues are highlighted in \Cref{fig:uv_frame_distortion}.

Since the aforementioned issues stem from the projection, transformation and interpolation of the UV appearance frame, masking can be applied based on the relative angle between the facial surface and the camera. Hence, we propose conditioning the UV appearance frame based on the relative orientation of the facial surface to the camera, to remove regions with re-projected and distorted appearance.

\noindent\textbf{Masking based on Relative Surface Orientation}
We begin by computing the relative angles for each mesh vertex, $\theta_{v} \in [-180^{\circ},180^{\circ})$, based on the cosine formula for the dot product between $\vec{n}_{v}$ and $\vec{n}_{cam}$. We compute $\vec{n}_{v}$ as the average of the triangle normal's which share that vertex, the triangle normal's are computed from the cross-product of two edge vectors.

Next, we apply the previously defined piece-wise affine transformations to transform the vertices from XY image-space into UV texture-space. Subsequently, spatial bi-linear interpolation of $\theta_{v}$ in the UV texture-space is performed to estimate the relative angle across the frame. This results in the representation of the relative surface angle frame in UV texture-space, denoted $\Theta_{UV}$, as shown in \Cref{fig:method/pipeline}.

\begin{figure}[H]
  \centering
  \begin{tikzpicture}[node distance=1.0cm and 1.0cm]
    \node (uv_frame_90) at (0,0) {\includegraphics[height=2.5cm]{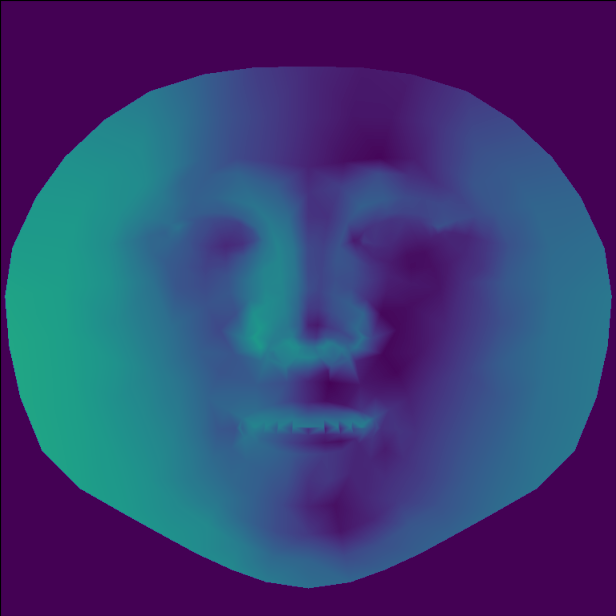}};
    
    \node (uv_frame_60) [right=of uv_frame_90, xshift=-1.0cm] {\includegraphics[height=2.5cm]{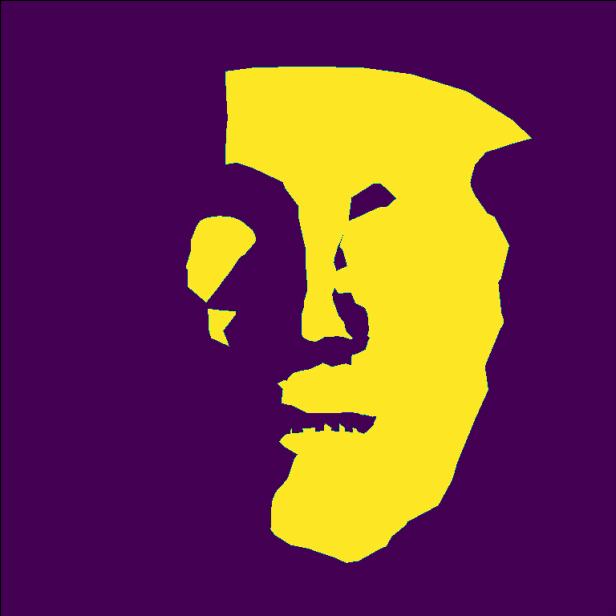}};

    \node (uv_frame_45) [right=of uv_frame_60, xshift=-1.0cm] {\includegraphics[height=2.5cm]{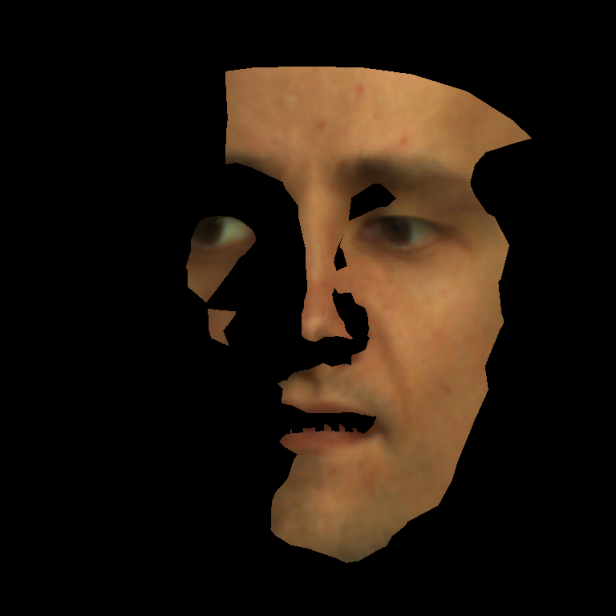}};
  
  \end{tikzpicture}

  \caption{Example of a computed UV angle frame $\Theta_{UV}$, the subsequent UV appearance mask for $\Theta_{UV} < 45^{\circ}$ and resultant masked UV appearance frame to be provided to the video-based model from PURE~\cite{noncontactvideobasedpulse-stricker-2014}.}%
  \label{fig:masked-uv-frame}
\end{figure}

We mask the UV facial frame based on $\Theta_{UV}$ to address the issues of re-projected and distorted appearance. Specifically, regions where $\Theta_{UV} \geq 90^{\circ}$ are masked to eliminate re-projected appearance. However, masking based on $\Theta_{UV}$ can remove frame information. Considering that appearance distortion is greater for $\Theta_{UV} \approx 90^{\circ}$, based on cross-dataset experimentation we apply masking where $\Theta_{UV} \geq 45^{\circ}$ to balance between removing information and distortion whilst eliminating re-projection, as illustrated in \Cref{fig:masked-uv-frame}.

\section{Experiments}
\label{sec:experiments}

\subsection{Experimental Setup}

Experiments for rPPG-based estimation of pulse rate (PR)~\cite{camerameasurementphysiological-mcduff-2023} are conducted on two publicly available datasets containing diverse motion scenarios: PURE~\cite{noncontactvideobasedpulse-stricker-2014} and MMPD~\cite{mmpdmultidomainmobile-tang-2023}.
\textbf{PURE}~\cite{noncontactvideobasedpulse-stricker-2014} is a small-scale dataset for facial rPPG estimation. It contains instantaneous PPG alongside 60 RGB videos for 10 subjects recorded with diverse head movement scenarios: steady, talking, slow/fast translation and small/medium rotation.
\textbf{MMPD}~\cite{mmpdmultidomainmobile-tang-2023} is a dataset for facial rPPG estimation under diverse conditions with comprehensive metadata. It contains instantaneous PPG at 30 Hz alongside 660 RGB videos at 30 FPS and 320 $\times$ 240 resolution for 33 subjects recorded with different head/body motions: stationary, rotation, talking and walking.

\subsection{Implementation Details}
\label{sec:implementation-details}

\noindent\textbf{Video-based Deep Learning Model:}
We adopt PhysNet~\cite{remotephotoplethysmographsignal-yu-2019} as our baseline video-based rPPG estimation model to evaluate the efficacy of our designed orientation-conditioned facial texture video representation.
Following the implementation and training details outlined in~\cite{rppgtoolboxdeepremote-liu-2023}, we trained PhysNet~\cite{remotephotoplethysmographsignal-yu-2019} with a batch size of 4 for 30 epochs using the Adam optimizer with a OneCycleLR scheduler using a maximum learning rate of 9e-3. We retained the model from the epoch with the lowest validation error for subsequent testing.

\noindent\textbf{Data Preparation:}
We perform cubic spline interpolation to adjust the sampling rate of the ground-truth rPPG signal to match the corresponding video sequence frame rate. Following the approach in~\cite{deepphysvideobasedphysiological-chen-2018, rppgtoolboxdeepremote-liu-2023}, we compute the first-order normalized signal difference as our ground-truth signal. 
We employ MediaPipe FaceMesh~\cite{mediapipeframeworkbuilding-lugaresi-2019} for 3D facial landmark detection with a confidence threshold of $0.45$. We linearly temporally interpolate the landmarks for up to three consecutive frames with missing landmarks.
We follow our proposed pipeline and compute the orientation-conditioned facial texture video representation with a frame size of 128 $\times$ 128 pixels. 
We then compute the first-order normalized frame difference~\cite{deepphysvideobasedphysiological-chen-2018, rppgtoolboxdeepremote-liu-2023} followed by pixel outlier-clipping and standardization~\cite{deepphysvideobasedphysiological-chen-2018} to serve as our input. We employ standardization per extracted window instead of per video as in~\cite{rppgtoolboxdeepremote-liu-2023}. 
Following~\cite{rppgtoolboxdeepremote-liu-2023}, we resize the frames to 72 $\times$ 72 pixels to facilitate equitable comparison with previously reported results given the known impact of frame size on model performance from~\cite{remotephotoplethysmographsignal-yu-2019}. 
We refer to PhysNet trained on this video representation as \textit{PhysNet-UV}. 
We do not employ any data augmentation techniques such as~\cite{paruchuri2024motion}, to isolate the focus of our study on the impact of the proposed orientation-conditioned facial texture video representation.

\noindent\textbf{Baseline Data Preparation:}
To enable comparison with standardized video preparation pipelines, we replace our orientation-conditioned facial texture video pipeline with static facial detection pipeline based on~\cite{rppgtoolboxdeepremote-liu-2023} as a baseline.
We leverage the previously obtained MediaPipe FaceMesh~\cite{mediapipeframeworkbuilding-lugaresi-2019} to derive a static bounding box based on the minimum and maximum bounds of the convex hull of the landmarks projected onto the XY-plane from the 0-th frame of a video sample. We scale the bounding box ($\times$1.5) with a fixed center and crop the video onto the resultant bounding box.
We resize the frames to 72 $\times$ 72 pixels and subsequently apply the first-order normalized frame difference, pixel outlier-clipping, and standardization operations as previously outlined.
We refer to PhysNet trained on this video representation as \textit{PhysNet-XY}.

\subsection{Evaluation}
\label{sec:eval}

\noindent\textbf{Pulse Rate Estimation:} 
To facilitate comparison with existing methods, we evaluate trained models using the downstream task of pulse rate (PR) estimation through a signal process of the estimated rPPG signal.
We begin by estimating the rPPG signal for the entire video, we apply the video-based model across all frames with a video using a sliding non-overlapping window and concatenate the results. This approach maximizes the number of samples and thus the frequency domain resolution from the Fast Fourier Transform (FFT), providing a higher resolution estimate of the PR. 
Following the standardized procedure outlined in~\cite{multitasktemporalshift-liu-2021, rppgtoolboxdeepremote-liu-2023} for estimating PR, we detrend the aggregated signal using~\cite{advanceddetrendingmethod-tarvainen-2002} and apply a 2nd-order Butterworth filter to the rPPG signal, with cut-off frequencies set to [0.75, 2.50] Hz to ensure equitable comparison with previously reported results~\cite{rppgtoolboxdeepremote-liu-2023}. This filtering step helps to remove noise and unwanted frequencies, enhancing the quality of the signal.
Then, we compute the estimated PR by identifying the dominant frequency within the power spectrum, which is computed using the FFT of the processed rPPG signal.
This process is applied to obtain both the predicted and ground-truth rPPG signals per video.

\noindent\textbf{Performance Metrics:}
Consistent with prior research~\cite{rppgtoolboxdeepremote-liu-2023}, we report commonly used performance metrics to evaluate model performance on PR estimation. 
These metrics include the mean absolute error (MAE) measured in Beats Per Minute (BPM), root mean square error (RMSE) (BPM) and Pearson's correlation coefficient ($r$) of the estimated PR per video across all videos within the test set to provide insight into the error, variance and correlation of the estimated PR.
We compute the signal-to-noise ratio (SNR)~\cite{robustpulserate-dehaan-2013} in decibels (dB) to provide insight into the frequency domain characteristics of the signals.
We additionally compute the standard error (SE) to provide a measure of the statistical accuracy of the different estimates in the full results.
The reporting for all performance metrics alongside standard errors are provided in the supplementary material in \Cref{sec:sup-intra-dataset} and \Cref{sec:sup-cross-dataset}. 
We refer readers to~\cite{rppgtoolboxdeepremote-liu-2023} for further details on these performance metrics.
Note we have used publicly available results, codes, and experimental settings for these error metrics and evaluation pipelines.

\subsection{Intra-dataset Testing}
\label{sec:experiments:intra}

\noindent\textbf{PR Estimation on PURE:} 
PhysNet-UV demonstrates a significant increase in the error compared to PhysNet-XY in intra-dataset testing on PURE~\cite{noncontactvideobasedpulse-stricker-2014}.
We strongly emphasize that intra-dataset testing on PURE~\cite{noncontactvideobasedpulse-stricker-2014} does not provide meaningful performance differentiation as is elaborated in \Cref{sec:discussion}, this highlights the importance of cross-dataset testing for meaningful evaluation.
We adopt a subject-exclusive 5-fold cross-validation training protocol for PURE~\cite{noncontactvideobasedpulse-stricker-2014}, the testing results are averaged across the folds to obtain the subject-independent performance.
We observe that samples from video \textit{Subject 7 - Talking} (S7-T) exhibit a large PR estimation error ($\approx$58 BPM) due to confounding frequency domain characteristics of the rPPG signal which the PR estimation pipeline fails to handle.
We report our results alongside existing works in \Cref{tab:intra/pure} and refer readers to \Cref{sec:sup-intra-dataset} in the supplementary material for the full results including the standard error, and results excluding \textit{Subject 7 - Talking}.

\begin{table}
    \centering
    \footnotesize
        \begin{tabularx}{0.905\linewidth}{l c c c}
            \toprule
            \multirow{2}{*}{Method} & MAE $\downarrow$ & RMSE $\downarrow$ & \multirow{2}{*}{\textit{r} $\uparrow$} \\
            & (BPM) & (BPM) \\
    
            \midrule
            CHROM~\cite{rppgtoolboxdeepremote-liu-2023} $\bigstar$ & 2.07 & 9.92 & 0.99 \\
            CHROM~\cite{rppgtoolboxdeepremote-liu-2023} $\bigstar$ & 5.77 & 14.93 & 0.81 \\ 
            2SR~\cite{novelalgorithmremote-wang-2016} $\bigstar$ & 2.44 & 3.06 & 0.98 \\
            POS~\cite{algorithmicprinciplesremote-wang-2017} $\bigstar$ & 3.14 & 10.57 & 0.95 \\
            POS~\cite{rppgtoolboxdeepremote-liu-2023} $\bigstar$ & 3.67 & 11.82 & 0.88 \\ 
            
            \midrule
            HR-CNN~\cite{visualheartrate-spetlik-2018} $\blacklozenge$ & 1.84 & 2.37 & 0.98 \\
            PhysNet~\cite{remotephotoplethysmographsignal-yu-2019} $\blacklozenge$ & 2.1 & 2.6 & 0.99 \\
            PhysNet+TFA+PFE~\cite{learningmotionrobustremote-li-2022} $\blacklozenge$ & 1.44 & 2.50 & - \\
            ETA-rPPGNet~\cite{etarppgneteffectivetimedomain-hu-2021} $\blacklozenge$ & \textbf{0.34} & \underline{0.77} & 0.99 \\
            
            \midrule
            Dual-GAN~\cite{dualganjointbvp-lu-2021} $\blacksquare$ & 0.82 & 1.31 & 0.99 \\
            Dual-TL~\cite{dualpathtokenlearnerremote-qian-2023} $\blacksquare$ & 0.36 & \textbf{0.68} & 0.99 \\
            rPPG-MAE~\cite{rppgmaeselfsupervisedpretraining-liu-2023} $\blacksquare$ & 0.40 & 0.92 & 0.99 \\

            \midrule
            PhysNet-XY (Excl. S7-T) $\blacklozenge$ & \underline{0.341} & 1.108 & \textbf{0.999} \\
            PhysNet-UV (Excl. S7-T) $\blacklozenge$ & 0.500 & 1.397 & \underline{0.998} \\
            
            \midrule
            \rowcolor{Maroon!10}PhysNet-XY (Ours) $\blacklozenge$ & 1.318 & 7.632 & 0.945 \\
            \rowcolor{Maroon!10}PhysNet-UV (Ours) $\blacklozenge$ & 1.639 & 8.919 & 0.924 \\
    
            & \color{BrickRed}\textbf{+24.3\%} & \color{BrickRed}\textbf{+28.7\%} & \color{BrickRed}\textbf{-2.2\%} \\
            
            \bottomrule
            
    \multicolumn{4}{p{200pt}}
    {
    $\bigstar$ Signal Processing; 
    $\blacklozenge$ Video-based Deep Learning; 
    $\blacksquare$ STMap-based Deep Learning;
    }
        \end{tabularx}

    \medskip

    \vspace{-8pt}
    \caption{Intra-dataset subject-independent performance on PURE~\cite{noncontactvideobasedpulse-stricker-2014}. Best results are marked in \textbf{bold} and second best in \underline{underline}.} 
    \label{tab:intra/pure}    
\vspace{-8pt}    
\end{table}

\subsection{Cross-dataset Testing}
\label{sec:experiments:cross}

\noindent\textbf{PR Estimation on MMPD:} 
PhysNet-UV demonstrates a significant 18.2\% reduction in the MAE (BPM) compared to PhysNet-XY in cross-dataset testing on MMPD~\cite{mmpdmultidomainmobile-tang-2023}, validating the efficacy of our proposed orientation-conditioned facial texture representation for enhancing the performance of facial rPPG estimation methods.
We evaluate the generalization performance of the PhysNet-XY and PhysNet-UV models trained on PURE~\cite{robustpulserate-dehaan-2013} by conducting cross-dataset testing on MMPD~\cite{mmpdmultidomainmobile-tang-2023} across all folds from \Cref{sec:experiments:intra}. We average the results across the folds to obtain the subject-independent performance.
We report our results in \Cref{tab:cross/pure-mmpd} alongside existing results on MMPD~\cite{mmpdmultidomainmobile-tang-2023} reported by~\cite{rppgtoolboxdeepremote-liu-2023}.
PhysNet-UV outperforms both the previous state-of-the-art and all deep-learning based approaches in terms of both MAE (BPM) and $r$ as compared to other methods also trained on PURE~\cite{noncontactvideobasedpulse-stricker-2014}.
We refer reads to \Cref{sec:sup-cross-dataset} in the supplementary material for the full results.

\begin{table}
    \centering

    \footnotesize
        \begin{tabularx}{0.895\linewidth}{l c c c c}
            \toprule
            \multirow{2}{*}{Method} & MAE $\downarrow$ & RMSE $\downarrow$ & \multirow{2}{*}{\textit{r} $\uparrow$} \\
            & (BPM) & (BPM) & \\
    
            \midrule
            CHROM~\cite{rppgtoolboxdeepremote-liu-2023} $\bigstar$ & 13.66 & \underline{18.76} & 0.08 \\
            POS~\cite{rppgtoolboxdeepremote-liu-2023} $\bigstar$ & \underline{12.36} & \textbf{17.71} & 0.18 \\
            
            \midrule
            DeepPhys~\cite{rppgtoolboxdeepremote-liu-2023} $\blacklozenge$ & 16.92 & 24.61 & 0.05 \\
            PhysNet~\cite{rppgtoolboxdeepremote-liu-2023} $\blacklozenge$ & 13.93 & 20.29 & 0.17 \\
            TS-CAN~\cite{rppgtoolboxdeepremote-liu-2023} $\blacklozenge$ & 13.93 & 21.61 & \underline{0.20} \\
            PhysFormer~\cite{rppgtoolboxdeepremote-liu-2023} $\blacklozenge$ & 14.57 & 20.71 & 0.15 \\
            EfficientPhys-C~\cite{rppgtoolboxdeepremote-liu-2023} $\blacklozenge$ & 14.03 & 21.62 & 0.17 \\
            
            \midrule
            \rowcolor{Maroon!10}PhysNet-XY (Ours) $\blacklozenge$ & 14.905 & 22.542 & 0.155 \\
            \rowcolor{Maroon!10}PhysNet-UV (Ours) $\blacklozenge$ & \textbf{12.187} & 19.849 & \textbf{0.294} \\
    
            & \color{ForestGreen}\textbf{-18.2\%} & \color{ForestGreen}\textbf{-11.9\%} & \color{ForestGreen}\textbf{+89.8\%} \\
            
            \bottomrule
    
        \multicolumn{4}{p{200pt}}
        {
        $\bigstar$ Signal Processing; 
        $\blacklozenge$ Video-based Deep Learning; 
        $\blacksquare$ STMap-based Deep Learning;
        }
    
        \end{tabularx}

    \medskip
    \vspace{-8pt}
    \caption{Cross-dataset subject-independent performance on MMPD~\cite{mmpdmultidomainmobile-tang-2023} trained on PURE~\cite{noncontactvideobasedpulse-stricker-2014}. Best results are marked in \textbf{bold} and second best in \underline{underline}.}
    \label{tab:cross/pure-mmpd}
    
\vspace{-8pt}    
\end{table}

\noindent\textbf{Motion Analysis on MMPD:} 
PhysNet-UV displays significant performance improvements across all tested motion scenarios compared to PhysNet-XY, validating the efficacy of our proposed video representation for enhancing the motion robustness of existing video-based facial rPPG estimation methods.
We follow the same process described in \Cref{sec:experiments:intra} to obtain the subject-independent performance for motion analysis on MMPD~\cite{mmpdmultidomainmobile-tang-2023}.
We report the comparison between motion scenarios in \Cref{tab:cross/pure-mmpd-motion}.
We demonstrate significant improvements in scenarios with rigid subject motion. We observe a lower performance improvement for \textit{Talking}, which contains a combination of subtle subject rigid and non-rigid head movement. 
Despite significant improvements, we still observe poor performance for \textit{Walking} which contains significant relative motion between the subject and the camera, highlighting the difficulty of robust PR estimation in the presence of dynamic and unconstrained subject motion.

\begin{table}
    \centering
    
    \footnotesize
        \begin{tabularx}{0.923\linewidth}{l c c c c c}
            \toprule
            \multirow{2}{*}{Scenario} & \multirow{2}{*}{PhysNet} & MAE $\downarrow$ & RMSE $\downarrow$ & \multirow{2}{*}{\textit{r} $\uparrow$} & \\
            & & (BPM) & (BPM) & \\
            \midrule
            
            \multirow{2}{*}{Stationary} & XY & 7.501 & 14.364 & 0.394 \\
            & UV & 5.887 & 11.931 & 0.553 \\
            \rowcolor{gray!50} & & \color{ForestGreen}\textbf{-21.5\%} & \color{ForestGreen}\textbf{-16.9\%} & \color{ForestGreen}\textbf{+40.6\%} \\
            \midrule

            Stationary & XY & 17.281 & 26.447 & 0.113 \\
            (after exercise) & UV & 15.522 & 24.298 & 0.247 \\
            \rowcolor{gray!50} & & \color{ForestGreen}\textbf{-10.2\%} & \color{ForestGreen}\textbf{-8.1\%} & \color{ForestGreen}\textbf{+118.9\%} \\
            \midrule
    
            \multirow{2}{*}{Rotation} & XY & 12.251 & 17.579 & 0.173 \\
            & UV & 8.627 & 14.659 & 0.344 \\
            \rowcolor{gray!50} & & \color{ForestGreen}\textbf{-29.6\%} & \color{ForestGreen}\textbf{-16.6\%} & \color{ForestGreen}\textbf{+98.0\%} \\
            \midrule
    
            \multirow{2}{*}{Talking} & XY & 8.979 & 14.898 & 0.381 \\
            & UV & 8.308 & 14.519 & 0.377 \\
            \rowcolor{gray!50} & & \color{ForestGreen}\textbf{-7.5\%} & \color{ForestGreen}\textbf{-2.5\%} & \color{BrickRed}\textbf{-1.2\%} \\
            \midrule
    
            \multirow{2}{*}{Walking} & XY & 28.490 & 33.202 & 0.036 \\
            & UV & 22.565 & 28.463 & 0.070 \\
            \rowcolor{gray!50} & & \color{ForestGreen}\textbf{-20.8\%} & \color{ForestGreen}\textbf{-14.3\%} & \color{ForestGreen}\textbf{+96.7\%} \\
            
            \bottomrule
        \end{tabularx}
    \medskip

    \vspace{-8pt}
    \caption{Cross-dataset subject-independent performance on MMPD~\cite{mmpdmultidomainmobile-tang-2023} of PhysNet-XY and PhysNet-UV trained on PURE~\cite{noncontactvideobasedpulse-stricker-2014} across different motion scenarios.}    
    \label{tab:cross/pure-mmpd-motion}
\vspace{-8pt}
\end{table}

\subsection{Ablation Study}
\label{sec:experiments:ablation}

Given the limited insights gained from intra-dataset testing on PURE~\cite{noncontactvideobasedpulse-stricker-2014} and the relevance of generalization performance in real-world contexts, we employ cross-dataset testing to evaluate our ablations. 
We train on PURE~\cite{robustpulserate-dehaan-2013} and test on MMPD~\cite{mmpdmultidomainmobile-tang-2023} using the same protocol described in \Cref{sec:experiments:cross}.
We denote the sequence of first-order normalized frame difference, pixel outlier clipping, and standardization operations described in \Cref{sec:implementation-details} as $F_{D}$ for brevity. We also denote the UV transformation operation as $T_{UV}$.

\noindent\textbf{Impact of $\Theta_{UV}$:}
Masking the facial texture using the surface orientation provides an effective mechanism to mitigate the negative impact of re-projected and distorted information in the UV coordinate frame as a result of the UV transformation process.
In this ablation, we vary $\Theta_{UV}$ to remove increasing amounts of re-projected and distorted information.
We report the results in \Cref{tab:ablation/processing} and highlight PhysNet-XY and PhysNet-UV for easier comparison.
We demonstrate that balancing the removal of distortion with retaining training information by masking regions with $\Theta_{UV} \geq 45^{\circ}$ is necessary to optimize the performance of the proposed facial texture video representation.

\begin{table*}
    \centering
    \footnotesize
        \begin{tabularx}{0.748\textwidth}{l l c c c c c}
            \toprule
            \multirow{2}{*}{Video Processing Pipeline} & MAE $\downarrow$ & RMSE $\downarrow$ & \multirow{2}{*}{\textit{r} $\uparrow$} & SNR $\uparrow$ \\
            & (BPM) & (BPM) & & (dB) \\
            
            \midrule
            
            \rowcolor{Maroon!10} Crop$_{Static}$ ($\times$1.5-Box) + Resize + $F_{D}$ (\textbf{PhysNet-XY}) & 14.905 & 22.542 & 0.155 & -6.882 \\
    
            Crop$_{Static}$ ($\times$1.5-Box) + Segment + Resize + $F_{D}$ & 15.237 & 23.524 & 0.120 & \textbf{-6.053} \\

            Crop$_{Dynamic}$ ($\times$1.5-Box) + Pad$_{Square}$ + Resize + $F_{D}$ & 17.988 & 25.183 & 0.033 & \underline{-6.263} \\

            Crop$_{Dynamic}$ ($\times$1.5-Box) + Pad$_{Square}$ + Segment + Resize + $F_{D}$ & 14.683 & 22.563 & 0.138 & -6.553 \\
            
    
    
    

    

            \midrule

            $T_{UV}$ + $F_{D}$ + Resize & \underline{12.687} & \underline{20.454}  & 0.248 & -6.679 \\
    
            $T_{UV}$ + Mask ($\Theta_{UV} \geq 90^{\circ})$ + $F_{D}$ + Resize & 13.038 & 20.900 & 0.216 & -6.473 \\
    
            $T_{UV}$ + Mask ($\Theta_{UV} \geq 60^{\circ})$ + $F_{D}$ + Resize & 12.890 & 20.629 & 0.256 & -6.284 \\

            \rowcolor{Maroon!10} $T_{UV}$ + Mask ($\Theta_{UV} \geq 45^{\circ}$) + $F_{D}$ + Resize (\textbf{PhysNet-UV}) & \textbf{12.187} & \textbf{19.849} & \textbf{0.294} & -6.265 \\
    
            $T_{UV}$ + Mask ($\Theta_{UV} \geq 30^{\circ})$ + $F_{D}$ + Resize & 13.300 & 20.834 & \underline{0.277} & -6.496 \\
    
            \bottomrule
            
        \end{tabularx}
    \medskip
    \vspace{-8pt}
    \caption{Ablative results with different video processing pipelines, obtained from cross-dataset testing on MMPD~\cite{mmpdmultidomainmobile-tang-2023} of PhysNet trained on PURE~\cite{robustpulserate-dehaan-2013}. Best results are marked in \textbf{bold} and second best in \underline{underline}.}
    \label{tab:ablation/processing}
\vspace{-8pt}     
\end{table*}

\noindent\textbf{Impact of Video Processing:}
The proposed facial texture representation provides improved performance over both dynamic facial localization and/or segmentation, highlighting the efficacy of the UV coordinate mapping process.
In this ablation, we vary the video processing pipeline to evaluate the impact of both static and dynamic facial detection and/or segmentation.
We derive the bounding box and segmentation mask from the projected XY landmarks to ensure consistency across experiments.
We report the results in \Cref{tab:ablation/processing} and highlight PhysNet-XY for easier comparison.
We refer readers to \cref{sec:sup-cross-dataset} in the supplementary material for the full results, including additional ablations.

We demonstrate that employing dynamic segmentation degrades the performance over the baseline, in-line with~\cite{benefitdistractiondenoising-nowara-2021}.
We also observe that employing dynamic facial detection degrades the performance, we suspect that dynamic facial detection results in jitter in the cropped region, resulting in large pixel-level changes in $F_{D}$.
However, dynamic facial cropping with segmentation may remove the large pixel-level variations in the background, improving the performance over the baseline.
We note the performance impact that subtle changes to steps and operations in the video processing pipeline can have.

\section{Discussion}
\label{sec:discussion}
We have proposed to model the 3D surface of the face as a strategy to disentangle rigid and non-rigid subject motion from video, reducing the spatio-temporal feature variability in video due to subject motion. We leveraged the 3D to 2D correspondence of UV coordinate texture mapping to construct video frames which enhance the performance and motion robustness of existing video-based facial rPPG estimation methods.
Our method achieves a 18.2\% cross-dataset performance improvement using the proposed orientation-conditioned facial texture video representation as demonstrated in \Cref{tab:cross/pure-mmpd} over our baseline, which represents a commonly employed video processing pipeline.
We demonstrated significant generalization performance improvements of up to 29.6\% across a diverse range of motion scenarios in \Cref{tab:cross/pure-mmpd-motion}, further validating the efficacy of our proposed video representation to improve motion-robustness.
We highlighted the importance of mitigating the effects of re-projected and distorted facial texture in \Cref{tab:ablation/processing} through leveraging the surface orientation.
We demonstrated the advantages of UV coordinate mapping over both dynamic facial detection and segmentation, and the impact of subtle changes in the video processing pipeline.
Our proposed orientation-conditioned facial texture video representation provides a robust and explicit inductive bias for enhancing the motion robustness of existing video-based rPPG methods.

\noindent\textbf{Limitations:} 
Our proposed representation inherently introduces distortion through the UV coordinate texture mapping process. We mitigated this by masking the appearance frame based on the facial surface orientation.
We observed performance trade-offs when masking a significant amount of training information, resulting in degraded performance as shown in \Cref{tab:ablation/processing}.
Furthermore, our proposed method explicitly relies on accurate and consistent 3D facial landmark detection to compute the orientation-conditioned facial texture video representation. 
Within this work we did not explore the impact of different 3D facial reconstruction methods or the effects of noisy landmark detection.

We emphasize the importance of increasing the frequency domain resolution of the FFT for fine-grained PR estimation, this is critical in scenarios where we expect high performance from our models.
Our evaluation process provides a frequency domain resolution of $\approx$0.88 BPM in PURE~\cite{noncontactvideobasedpulse-stricker-2014} , indicating the accuracy of the trained model is below the resolution of the evaluation task.
Furthermore we observe significant variability in the PR estimation for certain subjects despite similar frequency domain characteristics, highlighting the difficulty in designing robust PR evaluation pipelines.
This concern is also highlighted by the cut-off frequencies of the band-pass filtering which correspond to 45-150 BPM, in high-intensity exercise we may reasonably except PR to exceed 150 BPM.
These issues significantly limit the meaningful insights and performance differentiation we can obtain.

\noindent\textbf{Future Work:} 
Future directions include exploring alternative ways to exploit the 3D facial structure for more performant and robust facial rPPG estimation.
Existing STMap-based~\cite{rppgmaeselfsupervisedpretraining-liu-2023} approaches commonly extract image regions, we expect these approaches to benefit from leveraging the 3D facial structure to extract temporally consistent surface regions for use in the STMap.
Exploring the use of our orientation-conditioned facial texture representation with more recent and powerful video-based architectures such as PhysFormer~\cite{physformerfacialvideobased-yu-2022} in this context may similarly provide performance benefits.

Further experimental work leveraging large-scale datasets such as VIPL-HR~\cite{viplhrmultimodaldatabase-niu-2019} is necessary to meaningfully evaluate the intra-dataset performance, given the limitations of training expressive deep-learning models on small-scale datasets such as PURE~\cite{noncontactvideobasedpulse-stricker-2014}.
Further evaluation of rPPG estimation methods in real-world scenarios is also necessary to identify and establish the performance limitations to be addressed moving forwards.

\noindent\textbf{Impact Statement:} 
Subject motion represents a significant challenge in camera-based facial rPPG estimation.
Addressing this challenge is crucial to unlocking the potential of camera-based remote physiological measurement, particularly in vital applications like telehealth.
However, it is important to acknowledge the risks of these technologies being used in an unethical manner, such as for covert measurement.
Therefore, responsible and ethical use and deployment in validated and regulated scenarios is critical for realizing the benefits of this impactful technology.

\section{Conclusion}
\label{sec:conclusion}
In this paper, we have demonstrated that our proposed orientation-conditioned facial texture video representation improves the performance and motion robustness of existing video-based facial rPPG estimation methods.
Using the proposed video representation, PhysNet~\cite{remotephotoplethysmographsignal-yu-2019} achieves a substantial 18.2\% overall performance improvement compared to our baseline, and demonstrates significant improvements of up to 29.6\% in all tested motion scenarios.
Despite the limitations associated with the proposed video representation, our results and further investigations \Cref{sec:experiments:ablation} underscore the strength of disentangling subject motion through UV coordinate mapping.
More generally, this represents an interesting direction for future research to explore explicitly leveraging the facial structure as a strong inductive bias for more robust facial rPPG estimation in challenging rigid and non-rigid subject motion scenarios.
We hope the work and insights demonstrated in this work will contribute towards ensuring reliable and performant facial rPPG estimation in real-world scenarios. \\

\noindent\textbf{Acknowledgments:} 
This work was supported by the MRFF Rapid Applied Research Translation grant (RARUR000158), CSIRO AI4M Minimising Antimicrobial Resistance Mission, and Australian Government Training Research Program (AGRTP) Scholarship. \\

\noindent\textbf{Compliance with Ethical Standards:} 
This study was performed in line with the principles of the Declaration of Helsinki. 
%
%
The experimental procedures involving human subjects described in this paper were approved by CSIRO Health and Medical Human Research Ethics Committee (CHMHREC) [ethics protocol 2022\_016\_LR] and the Australian National University Human Research Ethics Committee (ANU HREC) [ethics protocols 2023/403 and 2023/483].

{
    \small
    \bibliographystyle{ieeenat_fullname}
    \bibliography{uv-motion-robustness}
}
\clearpage
\setcounter{page}{1}
\maketitlesupplementary


\section{Supplementary Material}
\label{sec:sup}

\subsection{Data Visualization}
\label{sec:sup-data visualization}

We provide select video frames approved for publication from the PURE~\cite{robustpulserate-dehaan-2013} dataset after different video processing steps but excluding the normalized frame-difference, pixel outlier clipping and standardization steps ($F_{D}$) for visualization and comparison.

\begin{figure}[H]
    \centering
    \includegraphics[width=0.35\linewidth]{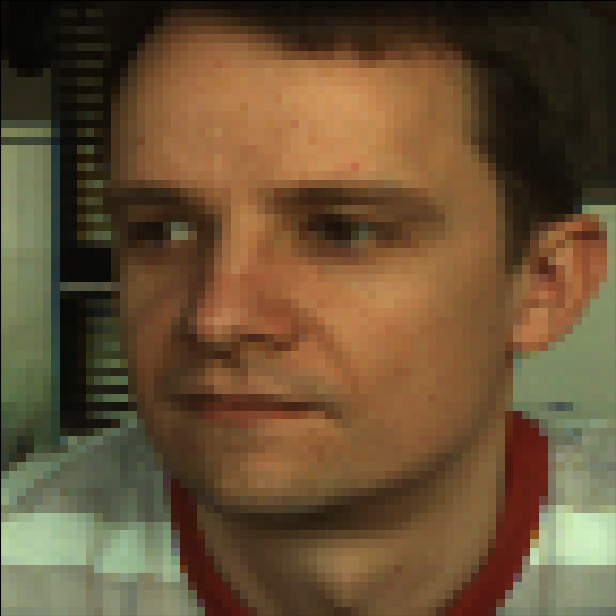}
    \caption{Frame of \textit{Subject 1} in scenario \textit{Medium Rotation} from PURE~\cite{robustpulserate-dehaan-2013} with static cropping ($\times$1.5-scale Box) applied.  Subsequent frames will use the same bounding box, hence the in-plane position of the face will vary due to subject motion.}
    \label{fig:pure-01-06-baseline}
\vspace{-16pt}
\end{figure}

\begin{figure}[H]
    \centering
    \includegraphics[width=0.35\linewidth]{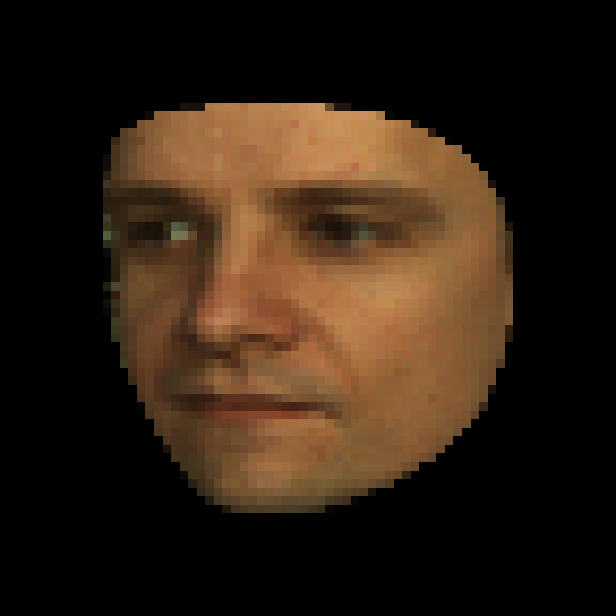}
    \caption{Frame of \textit{Subject 1} in scenario \textit{Medium Rotation} from PURE~\cite{robustpulserate-dehaan-2013} with static cropping ($\times$1.5-scale Box) and facial segmentation applied. Subsequent frames will use the same bounding box, hence the in-plane position of the face will vary due to subject motion.}
    \label{fig:pure-01-06-baseline_segment}
\vspace{-16pt}
\end{figure}

\begin{figure}[H]
    \centering
    \includegraphics[width=0.35\linewidth]{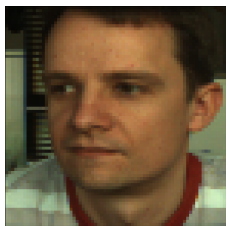}
    \caption{Frame of \textit{Subject 1} in scenario \textit{Medium Rotation} from PURE~\cite{robustpulserate-dehaan-2013} with dynamic cropping ($\times$1.5-scale Box) and square padding applied. However, subsequent frames will remain centered on the square padded and scaled facial region allowing for larger in-plane subject motion within a video sequence.}
    \label{fig:pure-01-06-baseline_dynamicCrop}
\end{figure}

\begin{figure}[H]
    \centering
    \includegraphics[width=0.35\linewidth]{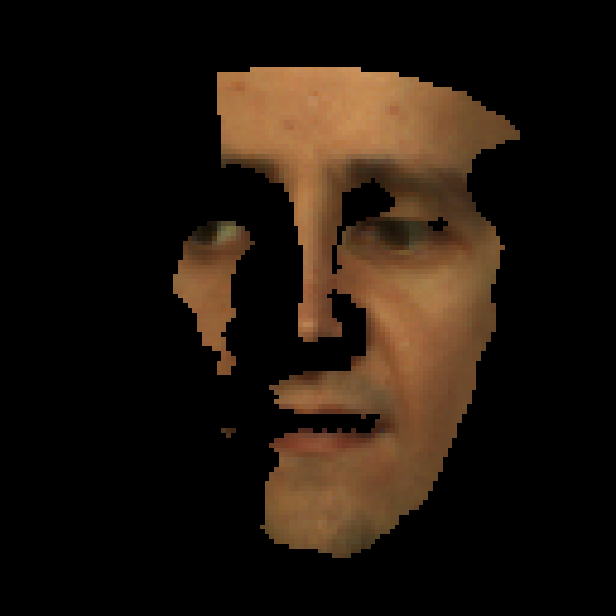}
    \caption{Frame of \textit{Subject 1} in scenario \textit{Medium Rotation} from PURE~\cite{robustpulserate-dehaan-2013} with UV transformation and masking ($\Theta \geq 45^{\circ}$) applied - the UV transformation process inherently dynamically localizes and segments the facial region. Subsequent frames will have the same structure with varying texture.}
    \label{fig:pure-01-06-baseline_uvMask45}
\end{figure}
\subsection{Intra-dataset Testing}
\label{sec:sup-intra-dataset}
In \Cref{tab:sup-intra-dataset} we report the full set of performance metrics referenced in \Cref{sec:eval} using the evaluation pipeline and metric implementations provided in~\cite{rppgtoolboxdeepremote-liu-2023} for intra-dataset testing on the PURE~\cite{noncontactvideobasedpulse-stricker-2014} dataset using subject-independent cross-validation. We obtain these results using the protocol described in \Cref{sec:experiments:intra}.
We report both the results including and excluding samples from \textit{Subject 7 - Talking} (S7-T) to provide insight into the evaluation variability.
We denote the sequence first-order normalized frame difference, pixel outlier clipping, and standardization operations as $F_{D}$ for brevity. We also denote the UV transformation operation as $T_{UV}$.

\begin{table*}
    \centering
    \footnotesize
    
        \begin{tabularx}{0.915\linewidth}{l c c c c}
            \toprule
            \multirow{2}{*}{Video Processing Pipeline} & MAE $\pm$ SE & RMSE $\pm$ SE & \multirow{2}{*}{\textit{r} $\pm$ SE} & \multirow{2}{*}{SNR $\pm$ SE} \\
            & (BPM) & (BPM) & & (dB) \\

            \midrule

            $F_{D}$ + Crop$_{Static}$ ($\times$1.5-Box) + Resize & 0.492 $\pm$ 0.172 & 1.408 $\pm$ 0.946 & 0.998 $\pm$ 0.008 & 10.721 $\pm$ 1.044 \\
            
            \midrule
    
            Crop$_{Static}$ ($\times$1.5-Box) + Resize + $F_{D}$ (\textbf{PhysNet-XY}) & 1.318 $\pm$ 0.979 & 7.632 $\pm$ 56.531 & 0.945 $\pm$ 0.043 & 11.061 $\pm$ 1.025 \\

            Crop$_{Static}$ ($\times$1.5-Box) + Resize + $F_{D}$ (Excl. S7-T) & 0.341 $\pm$ 0.138 & 1.108 $\pm$ 0.727 & 0.999 $\pm$ 0.007 & 11.457 $\pm$ 0.963 \\
    
            \midrule

            $T_{UV}$ + $F_{D}$ + Resize & 2.734 $\pm$ 1.510 & 11.918 $\pm$ 98.699 & 0.862 $\pm$ 0.067 & 11.546 $\pm$ 1.135 \\
    
            $T_{UV}$ + $F_{D}$ + Resize (Excl. S7-T) & 0.594 $\pm$ 0.217 & 1.739 $\pm$ 1.435 & 0.996 $\pm$ 0.011 & 12.228 $\pm$ 1.063 \\
    
            \midrule

            $T_{UV}$ + Mask ($\Theta_{UV} \geq 90^{\circ})$ + $F_{D}$ + Resize & 1.393 $\pm$ 0.938 & 7.338 $\pm$ 51.500 & 0.949 $\pm$ 0.042 & 12.011 $\pm$ 1.140 \\
    
            $T_{UV}$ + Mask ($\Theta_{UV} \geq 90^{\circ})$ + $F_{D}$ + Resize (Excl. S7-T) & 0.462 $\pm$ 0.171 & 1.381 $\pm$ 0.959 & 0.998 $\pm$ 0.008 & 12.470 $\pm$ 1.063 \\
    
            \midrule

            $T_{UV}$ + Mask ($\Theta_{UV} \geq 60^{\circ})$ + $F_{D}$ + Resize & 1.676 $\pm$ 1.316 & 10.243 $\pm$ 102.807 & 0.899 $\pm$ 0.058 & 12.211 $\pm$ 1.084 \\
    
            $T_{UV}$ + Mask ($\Theta_{UV} \geq 60^{\circ})$ + $F_{D}$ + Resize (Excl. S7-T) & 0.356 $\pm$ 0.139 & 1.114 $\pm$ 0.727 & 0.999 $\pm$ 0.007 & 12.617 $\pm$ 1.023 \\
    
            \midrule

            $T_{UV}$ + Mask ($\Theta_{UV} \geq 45^{\circ})$ + $F_{D}$ + Resize (\textbf{PhysNet-XY}) & 1.639 $\pm$ 1.141 & 8.919 $\pm$ 76.940 & 0.924 $\pm$ 0.051 & 11.842 $\pm$ 1.106 \\
    
            $T_{UV}$ + Mask ($\Theta_{UV} \geq 45^{\circ})$ + $F_{D}$ + Resize (Excl. S7-T) & 0.500 $\pm$ 0.171 & 1.397 $\pm$ 0.958 & 0.998 $\pm$ 0.008 & 12.159 $\pm$ 1.079 \\
    
            \midrule

            $T_{UV}$ + Mask ($\Theta_{UV} \geq 30^{\circ})$ + $F_{D}$ + Resize & 1.594 $\pm$ 1.113 & 8.693 $\pm$ 72.994 & 0.928 $\pm$ 0.049 & 11.486 $\pm$ 1.108 \\
    
            $T_{UV}$ + Mask ($\Theta_{UV} \geq 30^{\circ})$ + $F_{D}$ + Resize (Excl. S7-T) & 0.485 $\pm$ 0.172 & 1.397 $\pm$ 0.958 & 0.998 $\pm$ 0.008 & 11.817 $\pm$ 1.077 \\
            
            \bottomrule
    
        \end{tabularx}
    \medskip

    \caption{Intra-dataset subject-independent performance of PhysNet across different video processing pipelines on the PURE~\cite{noncontactvideobasedpulse-stricker-2014} dataset using averaged results across all folds from subject-independent cross-validation training on the PURE~\cite{robustpulserate-dehaan-2013} dataset.}
    
    \label{tab:sup-intra-dataset}
       
\end{table*}
\subsection{Cross-dataset Testing}
\label{sec:sup-cross-dataset}
In \Cref{tab:sup-cross-dataset} we report the full set of performance metrics referenced in \Cref{sec:eval} using the evaluation pipeline and metric implementations provided in~\cite{rppgtoolboxdeepremote-liu-2023}.
We obtained these results using the protocol described in \Cref{sec:experiments:cross}, we perform cross-dataset testing on the MMPD~\cite{mmpdmultidomainmobile-tang-2023} dataset using PhysNet models trained on the PURE~\cite{robustpulserate-dehaan-2013} dataset.
We report additional ablations for the operations applied after $T_{UV}$ to demonstrate the impact of the sequence of operations, and provide additional internally consistent comparisons.
We denote the sequence first-order normalized frame difference, pixel outlier clipping, and standardization operations as $F_{D}$ for brevity. We also denote the UV transformation operation as $T_{UV}$.

\begin{table*}
    \centering
    \footnotesize
        \begin{tabularx}{0.96\linewidth}{l c c c c}
            \toprule
            \multirow{2}{*}{Video Processing Pipeline} & MAE $\pm$ SE & RMSE $\pm$ SE & \multirow{2}{*}{\textit{r} $\pm$ SE} & \multirow{2}{*}{SNR $\pm$ SE} \\
            & (BPM) & (BPM) & & (dB) \\

            \midrule

            $F_{D}$ + Crop$_{Static}$ ($\times$1.5-Box) + Resize & 17.492 $\pm$ 0.307 & 24.827 $\pm$ 16.908 & 0.047 $\pm$ 0.017 & -6.225 $\pm$ 0.074 \\
            
            \midrule
            
            Crop$_{Static}$ ($\times$1.5-Box) + Resize + $F_{D}$ (\textbf{PhysNet-XY}) & 14.905 $\pm$ 0.295 & 22.542 $\pm$ 15.837 & 0.155 $\pm$ 0.017 & -6.882 $\pm$ 0.080 \\
    
            Crop$_{Static}$ ($\times$1.5-Box) + Segment + Resize + $F_{D}$ & 15.237 $\pm$ 0.312 & 23.524 $\pm$ 17.217 & 0.120 $\pm$ 0.017 & -6.053 $\pm$ 0.088 \\

            Crop$_{Dynamic}$ ($\times$1.5-Box) + Pad$_{Square}$ + Resize + $F_{D}$ & 17.988 $\pm$ 0.307 & 25.183 $\pm$ 16.488 & 0.033 $\pm$ 0.017 & -6.263 $\pm$ 0.072 \\

            Crop$_{Dynamic}$ ($\times$1.5-Box) + Pad$_{Square}$ + Segment + Resize + $F_{D}$ & 14.683 $\pm$ 0.298 & 22.563 $\pm$ 15.526 & 0.138 $\pm$ 0.017 & -6.553 $\pm$ 0.082 \\
            
            \midrule
    
            $T_{UV}$ + Resize + $F_{D}$ & 13.168 $\pm$ 0.285 & 21.014 $\pm$ 14.394 & 0.227 $\pm$ 0.017 & -6.606 $\pm$ 0.084 \\
    
            $T_{UV}$ + Mask ($\Theta_{UV} \geq 90^{\circ})$ + Resize + $F_{D}$ & 13.547 $\pm$ 0.288 & 21.391 $\pm$ 14.610 & 0.210 $\pm$ 0.017 & -6.644 $\pm$ 0.085 \\
    
            $T_{UV}$ + Mask ($\Theta_{UV} \geq 60^{\circ})$ + Resize + $F_{D}$ & 12.949 $\pm$ 0.284 & 20.840 $\pm$ 14.129 & 0.243 $\pm$ 0.017 & -6.305 $\pm$ 0.087 \\

            $T_{UV}$ + Mask ($\Theta_{UV} \geq 45^{\circ}$) + Resize + $F_{D}$ & 15.222 $\pm$ 0.302 & 23.072 $\pm$ 15.474 & 0.156 $\pm$ 0.017 & -6.105 $\pm$ 0.083 \\
    
            $T_{UV}$ + Mask ($\Theta_{UV} \geq 30^{\circ})$ + Resize + $F_{D}$ & 15.771 $\pm$ 0.298 & 23.285 $\pm$ 15.670 & 0.133 $\pm$ 0.017 & -6.643 $\pm$ 0.082 \\

            \midrule

            $T_{UV}$ + $F_{D}$ + Resize & 12.687 $\pm$ 0.280 & 20.454 $\pm$ 13.843 & 0.248 $\pm$ 0.017 & -6.679 $\pm$ 0.088 \\
    
            $T_{UV}$ + Mask ($\Theta_{UV} \geq 90^{\circ})$ + $F_{D}$ + Resize & 13.038 $\pm$ 0.285 & 20.900 $\pm$ 14.249 & 0.216 $\pm$ 0.017 & -6.473 $\pm$ 0.086 \\
    
            $T_{UV}$ + Mask ($\Theta_{UV} \geq 60^{\circ})$ + $F_{D}$ + Resize & 12.890 $\pm$ 0.280 & 20.629 $\pm$ 13.794 & 0.256 $\pm$ 0.017 & -6.284 $\pm$ 0.088 \\

            $T_{UV}$ + Mask ($\Theta_{UV} \geq 45^{\circ}$) + $F_{D}$ + Resize (\textbf{PhysNet-UV}) & 12.187 $\pm$ 0.273 & 19.849 $\pm$ 13.102 & 0.294 $\pm$ 0.017 & -6.265 $\pm$ 0.092 \\
    
            $T_{UV}$ + Mask ($\Theta_{UV} \geq 30^{\circ})$ + $F_{D}$ + Resize & 13.300 $\pm$ 0.279 & 20.834 $\pm$ 13.611 & 0.277 $\pm$ 0.017 & -6.496 $\pm$ 0.087 \\
            
            \bottomrule
            
        \end{tabularx}
        
    \medskip
    \caption{Cross-dataset performance of PhysNet across different video processing pipelines on the MMPD~\cite{noncontactvideobasedpulse-stricker-2014} dataset using averaged results across all folds from subject-independent cross-validation training on the PURE~\cite{noncontactvideobasedpulse-stricker-2014} dataset.}
    
    \label{tab:sup-cross-dataset}
    
\end{table*}

\end{document}